\documentclass[fleqn,10pt]{wlscirep}
\usepackage[utf8]{inputenc}
\usepackage[T1]{fontenc}
\usepackage{lineno}
\usepackage{graphicx}
\usepackage{comment}
\usepackage{float}
\usepackage{pdflscape}
\usepackage{adjustbox}

\title{Using machine learning to understand causal relationships between urban form and travel CO$_2$ emissions across continents}

\author[1,2,*]{Felix Wagner}
\author[2,1]{Florian Nachtigall}
\author[3]{Lukas Franken}
\author[2,1]{Nikola Milojevic-Dupont}
\author[1]{Nicolas Koch}
\author[6]{Rafael H. M. Pereira}
\author[2,4]{Jakob Runge}
\author[5]{Marta C. Gonzalez}
\author[1,2]{Felix Creutzig}
\affil[1]{Mercator Research Institute of Global Commons and Climate Change, Berlin, 10829, Germany}
\affil[2]{Technical University Berlin, Berlin, 10623, Germany}
\affil[3]{The University of Edinburgh, School of Engineering, Edinburgh, Scotland, United Kingdom}
\affil[4]{German Aerospace Center, Institute of Data Science, 07745 Jena, Germany}
\affil[5]{Department of Civil and Environmental Engineering, UC Berkeley, California, United States}
\affil[6]{Institute for Applied Economic Research, Brasília, Brazil}

\affil[*]{wagner@mcc-berlin.net}

\begin{abstract}
Climate change mitigation in urban mobility requires policies reconfiguring urban form to increase accessibility and facilitate low-carbon modes of transport. 
However, current policy research has insufficiently assessed urban form effects on car travel at three levels: (1) \textit{Causality} -- Can causality be established beyond theoretical and correlation-based analyses? (2) \textit{Generalizability} -- Do relationships hold across different cities and world regions? (3) \textit{Context specificity} -- How do relationships vary across neighborhoods of a city? 
Here, we address all three gaps via causal graph discovery and explainable machine learning to detect urban form effects on intra-city car travel, based on mobility data of six cities across three continents. 
We find significant causal effects of urban form on trip emissions and inter-feature effects, which had been neglected in previous work. Our results demonstrate that destination accessibility matters most overall, while low density and low connectivity also sharply increase CO$_2$ emissions. These general trends are similar across cities but we find idiosyncratic effects that can lead to substantially different recommendations. In more monocentric cities, we identify spatial corridors -- about 10--50 km from the city center -- where subcenter-oriented development is more relevant than increased access to the main center. Our work demonstrates a novel application of machine learning that enables new research addressing the needs of causality, generalizability, and contextual specificity for scaling evidence-based urban climate solutions.

\end{abstract}

\begin{document}

\flushbottom
\maketitle

\thispagestyle{empty}
\section*{Significance statement}
Car travel is one of the activities emitting the most greenhouse gas emissions in cities. Urban planning strategies can reduce urban car travel and thus mitigate climate change. How urban development impacts traveled distances is a long-standing problem that has been primarily studied via correlation-based studies in individual cities. Our study based on causal methods offers evidence that certain strategies, such as increasing accessibility to the city center, hold across six cities on three continents. We also show how to adapt strategies to the specifics of each city, thus offering a new approach of relevance to city governments, transportation planning researchers, and the climate change mitigation community.

\newpage
\section*{Introduction}
Cities are responsible for 70\% of the world's CO\textsubscript{2} emissions \cite{worldbank2021} and by 2050 nearly 70\% of the earth's population will live in urban areas \cite{worldbank2022}. These urban areas have a pivotal role in defining how humanity will solve the climate crisis. The largest CO\textsubscript{2} emitter in cities next to the building sector is urban transport, being responsible for 3 GtCO\textsubscript{2}-eq per year \cite{creutzig2016urban}. For promoting sustainable urban transport, altering physical infrastructure -- buildings, streets, and relevant points of interest, which together characterize urban form -- is widely accepted to be a key leverage point. Such infrastructure modifications are estimated to have stronger influence than personal or social factors \cite{javaid2020determinants}, as also emphasized by the IPCC \cite{creutzig2022demand}. Yet, it is currently unclear how to translate the IPCC's generic, national-level policies, such as more urban densification, into location-specific actions implemented by municipalities.

Three main bottlenecks hinder the uptake of urban planning initiatives that can effectively reduce urban car travel and induced CO\textsubscript{2} emissions: \textit{causality}, \textit{generalizability}, and \textit{context specificity}. 

First, urban planning decisions require an understanding of how urban form causes car travel emissions. A literature body spanning several decades has studied the impact of urban form on traveled distances, respectively described by the '\textit{6Ds of compact development}' and the metric `\textit{vehicle kilometers traveled}' (VKT). The 6Ds describe the destination accessibility, density, distance to transit, diversity, design, and demographic properties of a given location \cite{ewing2017does}. Most prior studies conclude that destination accessibility, described by the distance of a trip origin to the city center (or sub-centers) has the strongest effects on VKT \cite{naess2017d,ding2018applying, ewing2017does, WAGNER2022, dingchuan2014, hong2014built}. This feature is followed by density measures, such as population (or job) density \cite{hong2014built, ding2018applying, ZAHABI20151, nasri2012, ewing2017does, stevens2017does, WAGNER2022, dingchuan2014,LINDSEY2011}, which reduce VKT with higher densities. Socio-demographic variables, such as the average income of an area, are also found to have a stronger association, mostly with negative \cite{ding2018applying, hong2014built, ewingtian2015, ZAHABI20151} and few with positive \cite{WAGNER2022, Vance2007} correlations. Neighborhood design, and specifically the number and density of road intersections, reduces GHG emissions from car travel \cite{nasri2012, ewing2017does, stevens2017does, dingchuan2014, Vance2007}. Land use diversity, such as land use mix ratios, or proximity to transit, also reduces GHG emissions slightly \cite{ding2018applying, ewing2017does, stevens2017does, ZAHABI20151, Vance2007}. 

However, the \textit{causality} of these effects remains uncertain and only few studies have considered that some of the 6Ds might be dependent of each other \cite{naess_2015, chauhan2022} -- risking to bias effect estimates \cite{naess_2015} and causing skepticism on the general impact of urban form \cite{stevens2017does}. Causal dependencies across the 6Ds arise as features such as destination access influence car travel on the metropolitan scale, while other 6Ds operate on a neighborhood level. Hence, the effects of features like street design or density on travel are likely dependent on their location relative to the center \cite{naess2021} - a question insufficiently investigated by previous investigated by previous research \cite{ewing2017does, ewing2018testing, WAGNER2022, ding2018applying, ZAHABI20151, nasri2012, dingchuan2014}. Some mixed method approaches better reflect the causal mechanisms via (quasi-)longitudinal study designs, combining questionnaires and interviews of residents before and after urban form changes with geographical analysis \cite{SCHEINER2023100820, naess2017d, naess2021}. However, such approaches are cost-intensive, hardly representative for all locations of a city, and not scalable to many cities around the globe. Here, novel big data methods, such as causal graph discovery, can now approximate such dependencies based on observational data \cite{Runge2019,runge2023causal}. Recent examples in the context of urban transport include the analysis of using graph discovery methods to understand causal drivers of urban form on mode choice \cite{chauhan2022, monteiro2021, Tai-Yu2017} or subway ridership \cite{HUANG2022100341}.

Second, scaling city-specific recommendations requires an understanding of what may hold across certain cities -- \textit{generalizability} -- and what is specific to the local context. Yet, previous work on urban form and car mobility has mostly focused on one \cite{ding2018applying, ZAHABI20151, WAGNER2022, dingchuan2014, LINDSEY2011} or two cities \cite{naess2017d} or several of the same country \cite{ewing2017does, nasri2012, stevens2017does,ewingtian2015, THAO2020, Vance2007}. Using more diverse sets of cities from varying countries with different sizes, densities, and centralities could support distinguishing findings holding general or local relevance. 

Third, effective urban planning policies require location-specific recommendations that consider and leverage the unique spatial characteristics of a city -- \textit{context specificity}. Previous literature has progressed from single city-wide effect measures of the 6Ds \cite{ewing2017does, stevens2017does} to uncovering nonlinear behaviors of individual effects, such as \cite{ding2018applying}. However, only little work was able to assign urban form effects to specific locations of a city, limiting their applicability for neighborhood-specific policy advice and real-world implementation. Recently, explainable machine learning methods enable the investigation of effects at individual locations. A case study in Berlin, for example, demonstrated that location-specific analysis allows the identification of areas with disproportionately high travel emissions due to low accessibility\cite{WAGNER2022}.

Here, we address all three gaps and conduct a spatially explicit, cross-city analysis of the causal effects of urban form on travel-related CO\textsubscript{2} emissions across 6 six cities on three different continents. Our contributions are three-fold: (1) We establish a directed-acyclic graph (DAG) representing car distance traveled for commute and a subset of the 6Ds at the trip origin to develop a causal understanding of how urban form impacts VKT across all considered cities. (2) We analyze how individual features contribute to VKT and resulting trip emissions per city, using causal Shapley values derived from training a supervised machine learning model. (3) Finally, we demonstrate how analyzing location-specific causal effects of urban form on trip emissions enables us to discover new urban planning strategies.

\section*{Results}

\subsection*{Causal urban form effects partially confirm previous assumptions}
\begin{figure}[H]
\centering
\includegraphics[width=1\textwidth]{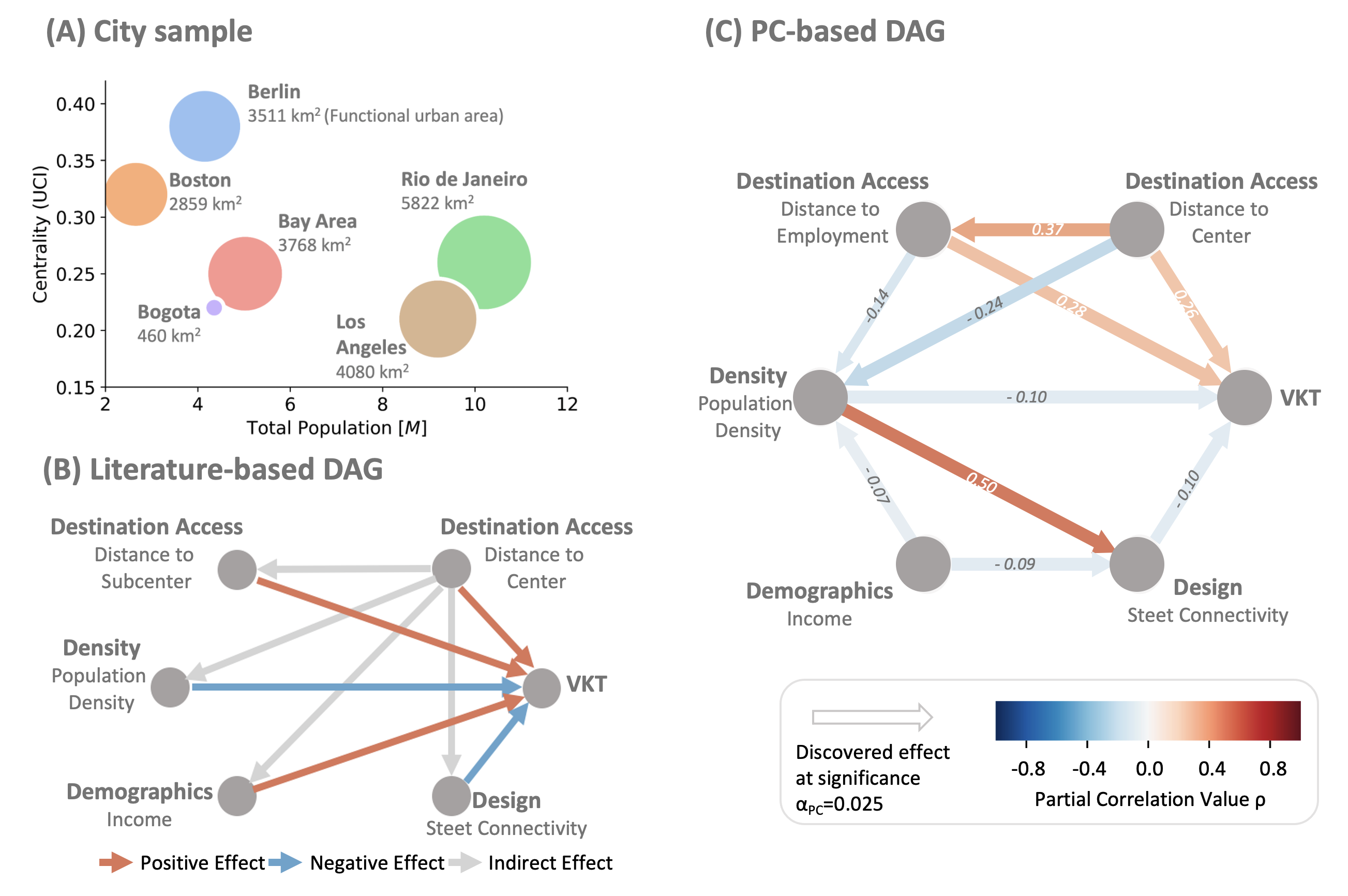}
\caption{\textbf{Comparison of causal graphs of the 6Ds of compact development on vehicle kilometers traveled (VKT) based on observational data across six cities.} \textit{Panel A:} Characterization of the cities Berlin (blue), Boston (orange), Los Angeles (brown), Bay Area (red), Rio de Janeiro (green), and Bogotá (purple) according to their centrality, measured by the urban centrality index on the y-axis (where values close to 0 indicate a more poly-centric distribution of jobs and values close to 1 indicate more mono-centric concentration), their total population count on the x-axis in million $[M]$ and functional urban area size (dot size in km$^2$). \textit{Panel B:} Causal directed-acyclic graph (DAG) displaying the relationships between Destination Accessibility, Density, Design, Demographics, and VKT from previous work. The coloring denotes a direct positive (red), direct negative (blue), and indirect (gray) effect. \textit{Panel C:} Causal DAG  based on the causal discovery with the PC algorithm using a robust partial correlation (Robust ParCorr) conditional independence test at a significance level $\alpha_{\rm PC} = 0.025$ (right). The arrow indicates the direction of the causal effect, while the coloring denotes a particular partial correlation value (see Methods, Causal Graph Discovery), indicating the causal strength of the effect between two nodes. No arrow between two nodes (e.g.~ between Demographics and VKT) indicates that no significant direct relationship was found between the two variables.
}
\label{fig1_dag}
\end{figure}

Using causal graph discovery, we find that destination accessibility, density, and design have a direct effect on VKT, while no significant direct effect is measured for income. In contrast with previous work, we observe inter-dependencies between several urban form features, enabling more nuanced urban planning decisions based on distinct cause-effect relationships. Fig.~\ref{fig1_dag} summarizes our findings.

To examine the causal relationship between urban form variables and VKT, we utilize data from six cities with varying area, population density, and centrality, measured by the urban centrality index (see Urban Centrality Index in Methods), as shown in Fig.~\ref{fig1_dag}, A. We apply the conditional independence test-based causal discovery framework~\cite{spirtes2000causation}, specifically the PC (Peter-Clark) algorithm~\cite{spirtes_glymour_1991,colombo2014order}, to an equally pooled sample of 1542 traffic assessment zones (TAZ) across all cities. The framework aims at detecting and orienting edges between nodes in a DAG using conditional independence tests. In our case, the nodes represent the target variable (VKT) and a subset of the 6Ds (see Feature Engineering in Methods for inclusion criteria). It also indicates a measure of causal strength of the dependency between two variables via the partial correlation value $\rho$. The discovered DAG can be interpreted as possibly causal for the analyzed data and constructed variables under certain assumptions (see Causal Graph Discovery in Methods for further details). Thus, our DAG allows for stronger statements towards causality than correlation-based studies \cite{runge2023causal}. We report the discovered DAG in Fig.~\ref{fig1_dag}, C. The DAG is compared with a graph that consists of a similar subset of the 6Ds, which was derived from previous relevant studies (refer to SI, Table \ref{table_feat_source_list}), as shown in Fig.~\ref{fig1_dag}, B.

Our PC-based DAG partly coincides with the literature-based DAG, but also exhibits noteworthy differences. Previous work has found all urban form features to have a direct effect on VKT. In previous studies, VKT typically correlates positively (red arrows) with distance to the center, distance to sub-centers and demographics (measured by mean income). VKT correlates negatively (blue arrows) with density and design. Additionally, some quasi-longitudinal studies recognized the different spatial scales of features and a resulting indirect link between distance to the center on all other urban from variables (gray arrow). In our study, we observe that lower distances to the main center and employment, higher population density, and street connectivity all have decreasing direct effects on VKT with partial correlation value $\rho$ ranging between $-$0.1 and 0.28. However, we cannot find a significant effect on VKT of demographics (measured by mean income). 

In addition, and importantly, we can measure for the first time relationships between urban form features. We find that distance to the main center strongly affects distance to employment ($\rho$ = 0.37) and population density ($\rho$ = $-$0,24), but we do not find significant indirect effects of distance to the center on other variables. We also find that distance to employment is negatively linked to population density ($\rho$ = $-$0.14) and population density is strongly positively linked to street connectivity ($\rho$ = 0.50). We observe that income negatively affects density ($\rho$ = $-$0.07) and connectivity ($\rho$ = $-$0.09), reflecting that higher-income households tend to be located in less dense areas. 


\subsection*{Trends generalize across cities but effect magnitudes differ}
\begin{figure}[H]
\centering
\includegraphics[width=1\textwidth]{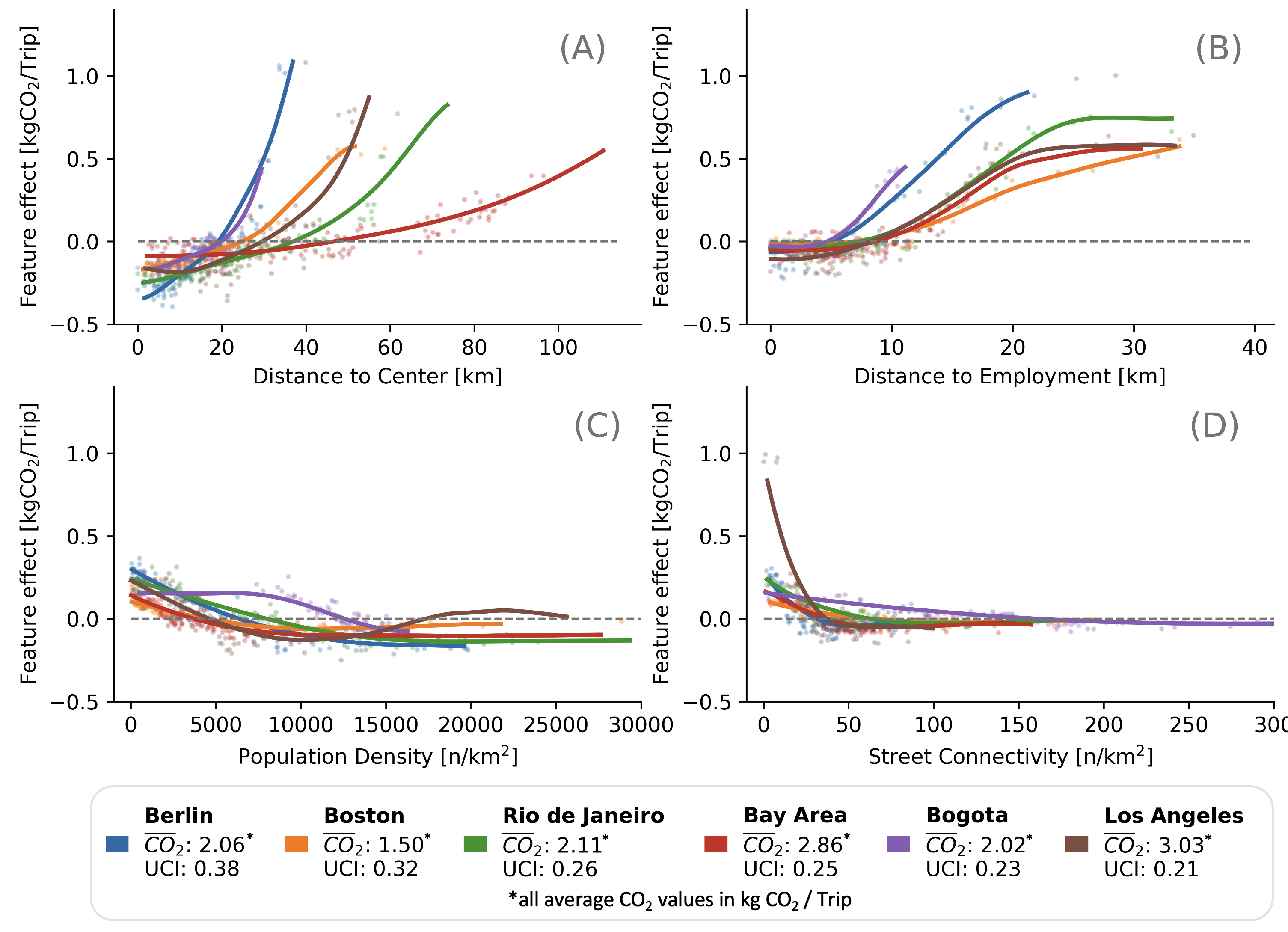}
\caption{\textbf{Comparison of causal Shapley effects on VKT for each significant urban form feature across cities.} Causal Shapley scatter plots are visualized for the features: distance to the center (A), distance to employment (B), population density (C), and street connectivity (D). Every dot of a given color represents a traffic assignment zone (TAZ) in a city. The solid lines represent a fit over all samples of a city. Feature effects are displayed relative to the city's average emission per trip in kgCO$_2$ (dotted line). Average trip emissions for each city ($\overline{CO}_2$) and urban centrality index (UCI) are provided in the legend to contextualize effect magnitudes.}
\label{fig2}
\end{figure}

When comparing urban form effects across cities (see Fig.~\ref{fig2}), we find that distance to the main center and distance to accessible jobs have stronger effects on trip emissions than density and connectivity. Yet, very low density and low connectivity increase emissions and should be avoided. While effects follow similar trends across cities, we measure differences in magnitudes that arise due to varying city sizes but also due to different spatial distributions of jobs and housing.

For the analysis of travel-related emissions, we convert observed travel distances to trip emissions (see SI, Location-specific emission factors). We first examine how well urban form explains trip emissions across cities. Here, we train a supervised, tree-based machine learning (ML) model in five cities and generalize to a sixth unseen city. We only consider features with a direct causal effect on the target to improve generalization performance. Then, we analyze the effect of individual urban form features on trip emissions at every location. For this, we calculate causal Shapley values \cite{heskes2020causal}, which incorporate the variable relationships from the learned DAG (for further details, see Methods, Modelling, Model Generalization and Interpretation).  

Overall, our ML model can capture the variation in trip emissions up to 84\% in Berlin, 62\% in Boston, 41\% in Rio de Janeiro, 26\% in the Bay Area, 51\% in Bogotá, and 21\% in Los Angeles (refer to SI, Table \ref{table_ml_results}). Trip emissions can be best explained in the mono-centric cities of our sample. 

The individual feature analysis via causal Shapley values reveals that distance to the center has the highest effect across the analyzed 6Ds. Shorter distances between trip origins and the center translate into lower CO$_2$ emissions. However, the feature distance to the center exerts non-linear effects that strongly increase between $22$ km (Berlin) and $55$ km (Bay Area) across all cities. The relative location to the center of a trip origin can affect emissions from $+500$ gCO$_2$ in Bogotá up to $+1100$ gCO$_2$ per trip in Berlin, respectively +24,7\% and +53\% relative to city average. Decreasing effects start from $-100$ gCO$_2$ (Bay Area) down to $-320$ gCO$_2$(Berlin) per trip, respectively -3.4\% and -15.5\% relative to city average. We find that in cities with longer distances to the center, the slope increases slower (compare the Bay Area and Bogotá). Yet, different magnitudes in increasing effects result not only from different city sizes but possibly also from where citizens who live far away from the center travel to. Here, more diverse destinations of trips starting very far away from the center tend to have smaller effect magnitudes (refer to SI, Fig.~\ref{si_fig2_A}). Additionally, the long tail of the Bay Area can be explained by the fact that we only define one center in downtown San Francisco. When defining four centers (including San Franciso, Oakland, Fremont and San Jose) the effect aligns closer with Boston and Los Angeles (compare SI, Fig.~\ref{si_fig2-sfo_4cbd}). 

\newpage

Distance to employment overall has slightly decreasing effects on VKT and related emissions for areas closer to job opportunities. This feature is nonetheless important, as we find that locations far away from jobs have strongly increasing effects on trip emissions beyond $6$--$7$ km in Bogotá and Berlin, and beyond $10$ km in Rio de Janeiro and US-based cities. This implies that depending on the location of accessible jobs, trip emissions can increase between $+550$ gCO$_2$ (Bogotá) and $+900$ gCO$_2$ (Berlin), respectively +27.2\% to +26,6\% relative to city average. 

For population density, we find that low population density corresponds to increasing effects between $+400$ gCO$_2$ (Berlin) and $+110$ gCO$_2$ (Boston), respectively +19.4\% and +7.3\% relative to city average. For higher densities, we find smaller, mixed effects between $-50$ gCO$_2$ and $-160$ gCO$_2$ ($-$3,3\% and $-$7.7\% relative to city average) per trip in Boston, the Bay Area, Rio de Janeiro, and Berlin. Two outlier cities display opposite trends, Bogotá and Los Angeles. In Bogotá, we find that the model allocates an increasing effect of $+200$ gCO$_2$ (+9.9\% relative to city average) for densities up to $+8000$ people per km$^2$. This might occur due to the overall very high densities in Bogotá, requiring the model to strongly extrapolate (compare SI, Table \ref{table_feature_distribution}). In Los Angeles, we measure slightly increasing effects for very high densities beyond $17$k people per km$^2$. Whether this is due to a lower model fit or particularly relevant in more polycentric cities requires further investigation with a focus on travel close to the center, as these trip origins mostly lie very close to downtown Los Angeles (refer to SI, Fig.~\ref{si_fig2_C}).

When analyzing street connectivity, we observe only small decreasing effects between $-30$ gCO$_2$ (Bogota) and $-140$ gCO$_2$ (Los Angeles) per trip for higher densities across all cities, respectively $-$1,5\% and $-$4.6\% relative to city average. Increasing effects appear below $20$ to $45$ intersections per km$^2$ for all cities apart from Bogotá. While the increasing effects stay below $+300$ gCO$_2$ per trip for most cities, in Los Angeles, we observe some outliers with strong increasing effects of up to $+1000$ gCO$_2$ (+33.3\%) per trip for very low densities. The outliers appear to be mostly very large TAZ with few intersections, which induce disproportionally low connectivity values, as analyzed in SI, Fig.~\ref{si_fig2_D}.

\subsection*{Which urban form effect matters most depends on specific locations within cities}
TAZ-level causal Shapley values also enable us to discover specific locations across all cities (see SI, Individual feature effects for effect maps), in which the lack of density, instead of the distance to the center increases trip emissions most. This points to particular planning opportunities for lowering commuting trip emissions of peripheral communities, based on sub-center development.

\begin{figure}[H]
\centering
\includegraphics[width=0.8\textwidth]{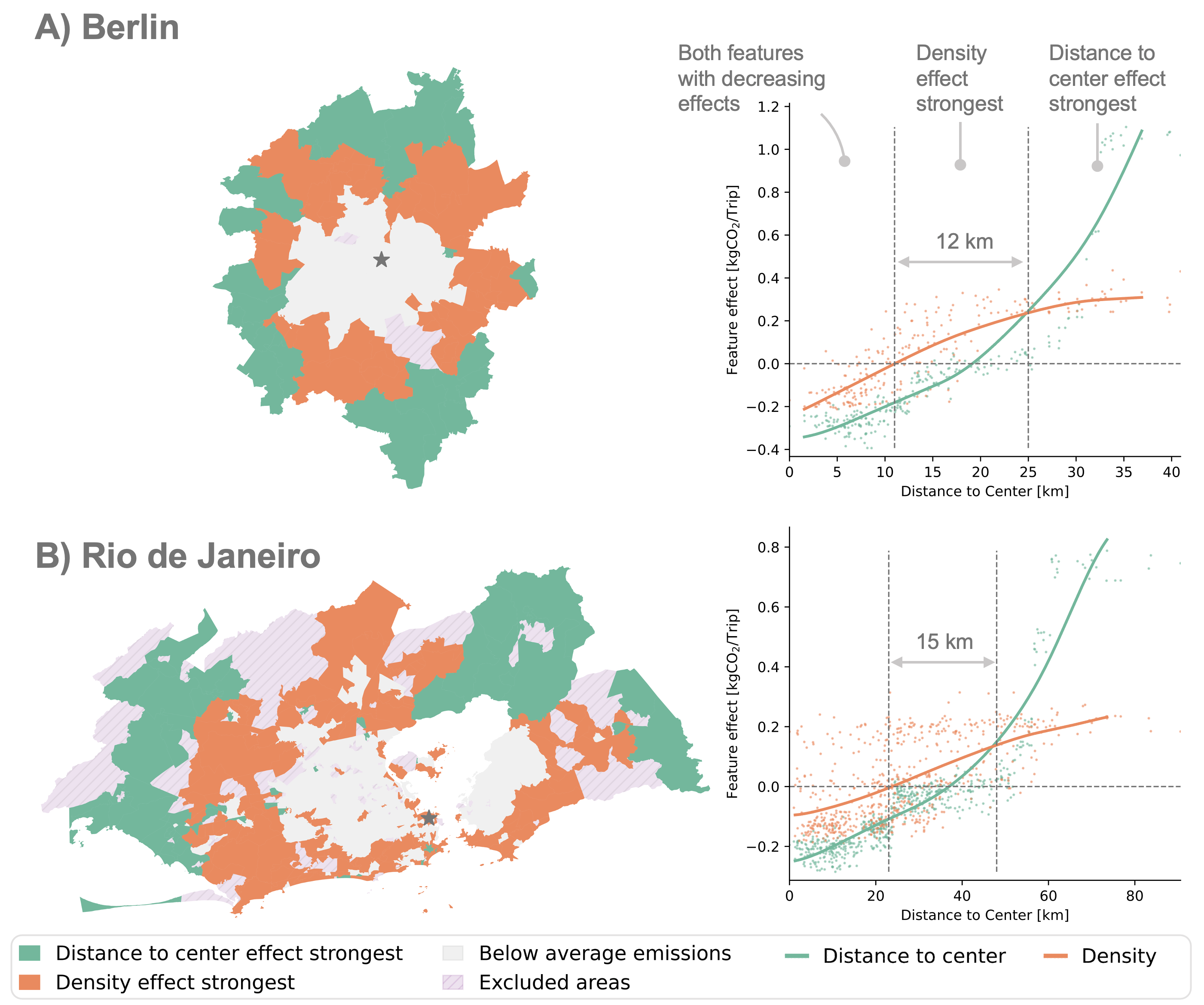}
\caption{\textbf{Location-specific urban form effects reveal corridors where density matters more than accessibility.} \textit{Left panels}: The map shows whether population density (orange) or distance to center (green) increase VKT most at Traffic Assingment Zonzes (TAZ) with emissions above the city average. Areas below average city emissions are coded in gray and purple areas correspond to TAZs excluded due to data cleaning. Results are shown for Berlin (A) and Rio de Janeiro (B). 
\textit{Right panels}: The scatter plots display for the respective city how the effects of both features evolve depending on the location of the TAZ relative to the center. Per color, every dot represents a traffic assignment zone (TAZ) in a city. The solid lines represent a fit over all samples of the city. Feature effects are displayed relative to the city's average emission per trip in kgCO$_2$ (horizontal dotted line). Vertical dotted lines reveal corridors where density increases emissions more than the distance to the center.
}
\label{fig3_spatial_shap_effects}
\end{figure}

For all locations with above-mean trip emissions, we examine if the measured causal Shapley effect of distance to the center or population density is larger, as shown in Fig.~\ref{fig3_spatial_shap_effects}. This reveals which of the two features are driving trip emissions the most and should be prioritized in low-carbon urban planning strategies.

Across all cities, we find specific locations where a lack of density is increasing emissions more strongly than the distance to the center. In monocentric cities, this difference unfolds around the center creating donut-shaped corridors, where first density, and after which it is the effect of distance to center that drives most trip emissions. In more polycentric cities, the effects are less spatially clustered, except in the very outskirts where the distance to the center always increases emissions most. We display in Fig.~\ref{fig3_spatial_shap_effects} the differences in Berlin and Rio de Janeiro, as examples for these two cases with high model accuracy (the results for all remaining cities can be found in the SI, Threshold effects of distance to center and population density, Fig.~\ref{si_fig3}). 

In Berlin and Rio de Janeiro, we find TAZ in which a lack of density has a higher increasing effect on emissions than the distance to the center. In Berlin, we find that close to the center, most locations exhibit below-average travel emissions (marked in gray) and both features have decreasing effects. As density decreases with longer distances from the center, it starts to have an increasing effect after $11$ km from the center (see the evolution of the orange line). At these locations, the effect of distance to the center is still negative and decreases emissions. At $23$ km, the distance to the center starts having a positive effect, and at $28$ km, it becomes larger than the effect of density. This creates a $12$ km buffer where density is more important than accessibility, and where the maximum differential between both effects is $300$ gCO$_2$ per trip, at $20$ km distance to the center. In Rio, we find similar yet less strong effect differences. Close to the center, we find that most areas have decreasing effects for both features. Yet, some areas in the proximity of the center also display increasing effects of density, like Alto da Boa Vista or the main harbor of the city. In Rio, density starts having increasing effects for most locations after $26$ km, while the distance to the center starts having an increasing effect only after $45$ km. At $51$ km, the effect of distance to center becomes higher than the effect of density, resulting in a buffer of $15$ km with a maximum effect differential of $150$ gCO$_2$.

\section*{Discussion}

\subsection*{Improving accessibility, avoiding low density and low connectivity}
By considering relationships between urban form and trip emissions via causal graph discovery and causal Shapley values, we find that accessibility of the center is most important for reducing trip emissions. We also find that low density and low connectivity should be avoided as they strongly increase emissions. 

The importance of high accessibility implies that new housing should be preferably located as close to the center as possible (compare decreasing effects in Fig.~\ref{fig2}, A). This recommendation is valid for all analyzed cities. With greater distances to the center, trip emissions increase disproportionally and alternative strategies become more important. In more mono-centric cities, we find city-specific corridors between 10--50km located between the inner city and the outskirts, where low-carbon planning strategies should prioritize local density over access to the center. In these areas low density is driving high trip emissions even more than low access. Here, creating secondary urban centers that increase density could help to reduce trip emissions for citizens who cannot or do not want to live closer to the center. Beyond observed corridors, the lack of access to the main center becomes so strong that car trips should be avoided if possible. While the avoidance of car trips reduces emissions everywhere in the city, it is most effective, yet most difficult, in the very outskirts. Here, possible strategies to reduce individual car travel include transit-oriented development (initiating mode shifts), ride-sharing initiatives (increasing vehicle occupancy), and telecommuting (avoiding trips). 

We also find across cities all that long distances to jobs increase emissions and mostly affect peripheral locations (compare increasing effects in Fig.~\ref{fig2}, B). While generating additional employment opportunities at the outskirts might alleviate those effects, they can come at the risk of inducing new car travel, as the city outskirts are less accessible via public transport and less congested, thus particularly prone to car usage. Hence, additional measures, such as increased public transport accessibility or ride-pooling offers should accompany such planning considerations.

\subsection*{Location-specific differences call for spatially-explicit analyses}
Our work emphasizes an existing bottleneck for climate change mitigation in cities, where municipalities must implement sustainable planning policies, yet lack the tools to adequately prioritize strategies and locations. Big data with high spatial resolution have a clear potential to fill this gap -- as our work demonstrates. We show how generic urban planning strategies like `compact development' can be translated into city-specific radii (compare where effects lines cross the zero line in Fig.~\ref{fig2}, A) and how different urban form effects can even be compared and prioritized at specific locations (see Fig.~\ref{fig3_spatial_shap_effects}).

The pressing climate crisis requires solutions to be modeled in more than just six cities. We call for more comparative research that can further discover possible generalizable trends. This research would support generating city typologies where recommendations can be replicated, similar to \cite{creutzig2015}. At the same time, it would specify how such trends should be modulated within cities according to idiosyncratic urban form effects. However, while open data on urban form has increased dramatically \cite{eubucco_2023, bing2022}, scaling the framework is constrained by the scarce availability of open and harmonized data on travel patterns, urban functions and socio-demographics at high resolution. Our results show the importance of looking are more cities: indeed, trip emissions can be explained well in more monocentric cities, like Berlin or Boston, with an accuracy of up to 81\%; yet, in other more poly-centric cities, like Los Angeles or the Bay Area, we get less clear results. This implies that there must exist other relevant factors. Such factors could include access to sub-centers of specific functions (such as work, commercial, leisure) or car connectivity between sub-centers. Those could not be modeled here mostly due to a lack of consistent available data across all cities. 

In addition, more work covering cities from the Global South and in different developing stages is primordial. This is necessary as, for example, cities with disproportionately high densities such as Bogotá require the model to extrapolate if little comparable cities are available for pattern recognition (see Fig.~\ref{fig2}, C). With more cities from the Global South such density effects could be more robustly assessed and may inspire novel transport strategies. Relevant strategies to reduce car commute emissions could include the installation of shared ride pooling services in peripheries that often struggle to be economically viable in low-dense, low-demand suburban areas in the Global North \cite{schlenther2023}. Similarly, further comparisons between mature and fast-growing cities would broaden the existing set of recommendations, as outlined in the Transport Chapter of the 6th IPCC assessment report \cite{ar6wg3_2022}. This would enable research to better specify differences in their specific planning needs, such as re-think strategies in mature cities, like the 15-minute concept, versus the avoidance of lock-in decisions in fast-growing cities.

\subsection*{Steering low-carbon urban transformation with causal science}
Via causal graph discovery, we demonstrate that urban form features are not independent, but rather interconnected. This provides empirical evidence that confirms previous hypotheses made in the literature about possible causal relationships between urban form and travel\cite{SCHEINER2023100820, naess_2019}. Our findings highlight that changes to one urban form dimension, such as density, are interdependent with others, such as the distance to the center. We strongly believe that reflecting such dependencies in analyses should become standard to ensure effective prioritization of planning efforts. 

Yet, we acknowledge that while the graph discovery minimizes required assumptions about if and how different urban form features are interlinked, it comes at the cost of other assumptions, such as that variable relationships are continuous and linear. These are made very explicit in Methods, Causal Graph Discovery. Yet, the obtained graph is influenced by them and results should be interpreted considering them (see SI, Causal Graph Discovery, Assumptions). In comparison to the existing (quasi-)longitudinal, mixed methods study design, our approach is representative of all locations of a city and scalable across continents. This enables us to compare cities within and across. Yet, our approach cannot capture all relevant dimensions. For example, we only proxy residential self-selection effects via income and find no significant direct effect when modeled across all cities. 
While influence on trip emissions has been found to be smaller than of urban form \cite{ding2018applying}, representing socio-demographic and attitude-based-residential-self selection most accurately would require additional mixed-method study designs, like \cite{SCHEINER2023100820, naess_2019}, which could incorporate our findings and further validate it for a local context.

\section*{Conclusion}
Our findings demonstrate the importance of causal and spatially explicit urban form analysis to derive a location-specific but generalizable understanding of climate change action. 
Across cities, we find that: (1) Distance to the center and to jobs reduces urban travel emissions most, followed by increased density and better street connectivity. This implies that center densification within city-specific radii is most effective for low-carbon urban travel. (2) Subcenter-oriented development in suburban spatial corridors between about 10-30km from the city center could further reduce emissions, especially in monocentric cities, like Berlin or Boston. (3) For the more poly-centric urban areas, like Los Angeles and the Bay Area, the chosen urban form features can only partially explain trip emissions, requiring additional climate policies.

This work is a starting point for future work at the intersection of big spatial data, causality, and climate change research -- highly relevant for urban planners, municipalities, and researchers alike. Our results demonstrate that big data approaches help to translate prevalent national-scale scenarios for climate change mitigation from the IPCC to local-level recommendations. 
Analyzing urban form relationships via causal graph discovery makes system dependencies explicit, revealing where interventions may be most effective.
The analysis via causal Shapley values allows to incorporate causal system relationships, and conjointly enables us to visualize effects spatially in maps where hotspots for actions can be identified. The resulting insights and visualizations are of high relevance for interdisciplinary urban design and citizen participation throughout the planning process. For researchers, the obtained results open up new opportunities, such as modeling urban form interventions via causal inference methods which can utilise the discovered DAG. Prospective modeling of the (treatment) effects of existing and future urban planning policies will improve the quality of evidence-based policy-making by a substantial margin. More broadly, as cities are complex systems \cite{BAI201669} our approach can initiate an integrated, systems-thinking approach to urban planning, as proposed by \cite{xuemaibai2023}, allowing municipalities and urban planners to better understand cause-effect mechanisms in spatially highly contextualized settings. 

\newpage
\section*{Methods}

\subsection*{Data}
\paragraph{Datasets}
For mobility data, we use various data sources from previous peer-reviewed studies. In Boston, Los Angeles, the Bay Area, Rio de Janeiro, and Bogotá, we use origin-destination tables for traffic demand based on call detail records (CDR) aggregated on TAZ as used in \cite{Ambühl2023, Colak2016, jiang2016, yanyan2017, Florez2017}. In Berlin, we utilize origin-destination tables based on GPS signals of navigation systems from various sources, including connected cars, floating cars and commercial car fleets, cleaned and calibrated as in a previous study \cite{WAGNER2022}. 

For all cities, urban form data is extracted from OpenStreetMap (OSM) \cite{osm2021}, while center definitions as well as population density, income and employment data are collected from region-specific sources as no harmonized data source could be obtained. The definition of centers are derived from Google Maps \cite{gmaps}, similar to \cite{Xu2023}. For the US cities, population count and household income data from the United States Census Bureau \cite{us_census} as of 2013 are used. For Rio de Janeiro, population count and average household income per capita data are obtained from the Access to Opportunities Project (AOP) for the year 2010 \cite{aop_brazil}. The data are aggregated by H3 hexagonal cells at a resolution of $9$. In Bogotá, population data is adopted from \cite{meta2017} and socioeconomic stratification on the block level as developed in \cite{Florez2017} is used as a proxy for income. In Berlin, population data on a 100mx100m grid is derived from \cite{zensus2011} and filled up at the boundaries with population density data from \cite{meta2017}. Income data on the household level is obtained from the commercial provider AXCIOM as used in \cite{WAGNER2022} and filled up with data from \cite{zeit2022} at the boundaries.
Employment data, defined by number of jobs per TAZ, is derived from the Smart Location Database \cite{epa_us} for the US cities and in Rio de Janeiro from the AOP project \cite{aop_brazil}. In Berlin and Bogotá, no open employment data could be obtained and employment locations are inferred from all work destinations derived from the SrV mobility survey 2017 \cite{srv2018} and Bogotá mobility survey 2015 respectively \cite{movilidad_Bogota}.


\paragraph{Cleaning}
From all mobility datasets only trips in the morning hours between $6$ and $10$ am that start and end within the boundary of a city are considered. For the boundary definition, we adopt the functional urban area definition \cite{eu_oecd_fua} as this recognizes the total city area and its surrounding commuting zone. At the outskirts, we consider all TAZ which overlap at least 50\% with the boundary. For Rio de Janeiro, we adopt the official boundary of Rio's 'urban concentration area' defined by Brazil's statistics Office \cite{ibge_rio_fur} as recommended by local knowledge of one of the manuscript's co-authors as it better reflects Rio de Janeiro's commuting catchment area.

The mobility data from the US cities, Rio de Janeiro and Bogotá are further cleaned and calibrated as in previous studies, which includes assigning mean number of trips from CDR records to TAZ. For Berlin, the same cleaning procedure as in \cite{WAGNER2022} is applied and individual trip origins and destinations are aggregated to zip codes (in the following we adopt the wording TAZ across all cities to describe the spatial unit of analysis) to match the other data sources. Additionally, only TAZ are considered that have at least ten trip origins to construct a meaningful average and TAZ with an airport are removed from the sample, as this study focuses on commuting.

\paragraph{Target variable -- VKT and Trip emissions }
To derive travel emissions, we transform mean travel distances to emissions using location specific emissions factors (refer to SI, Location specific emission factors). For the US cities, Rio de Janeiro, and Bogotá the mean travel distances per TAZ are calculated by creating uniformly random samples of trip origins and destinations on the road network. For each TAZ, we sample the number of points starting and ending within the TAZ's boundaries. To reduce errors we only sample onto residential or tertiary roads, as we assume that no trips start on highways, primary or secondary roads. To calculate travel distances, we calculate the shortest path along the street network from origin to destination and average travel distances per TAZ. In Berlin, we similarly average the travel distances of all trips starting within a TAZ.

\paragraph{Feature Engineering}
We develop five urban form features across all cities to represent four of the 6Ds. We define the features based on what has been used most frequently across previous work and for which data is available across all cities, compare Table \ref{table_feat_source_list}. Our features represent destination accessibility, measured by distance to center and employment, density, measured as population density, demographics, measured by income, and design, represented via street network connectivity. We choose to exclude the diversity dimension, due to data restrictions and as previous studies only associated with it (if at all) only small effects in the context of car travel and as it is more relevant for promoting active travel \cite{ewing2017does, stevens2017does}. As everyone in our study has already decided to use the car, we also exclude distance to transit, which primarily influences mode choice, but hardly car travel distance \cite{javaid2020determinants, aston2021exploring}. Table \ref{feat_list} summarises the features.

We acknowledge that previous work has emphasized the significance of addressing residential self-selection by including socio-demographic factors and travel attitudes \cite{naess_2019, handy2005correlation, devos2021indirect}. Since detailed data on travel-related attitudes is not available across all cities, we can only control for the effects of residential self-selection that can be proxied via the socio-demographic variable income, similar to \cite{ding2018applying, WAGNER2022}.

\begin{table}[H]
\small
\centering
\begin{tabular}{ lll }
\textbf{D-Variable} &\textbf{Feature} & \textbf{Description} \\
\hline
Destination Acessibility & Distance to city center & Distance from TAZ center to main city center. \\
Destination Acessibility & Distance to employment & Weighted average distance from TAZ center to 1\% of all jobs. \\[0.1cm]
Density & Population density & Number of inhabitants divided by area of TAZ. \\[0.1cm]
Demographics & Income  &  Average household income per TAZ. \\ [0.1cm]
Design & Street connectivity & Number of intersections divided by area of TAZ. \\ [0.1cm]
\end{tabular}
\caption{\textbf{Predictive features of urban form based on the 6Ds of compact development}}
\label{feat_list}
\end{table}

\subsection*{Modelling}
\paragraph{Urban Centrality Index}
We measure the centrality of an urban area adopting the urban centrality index \cite{uci_2013}. It controls for differences in areas and shape of a city and assigns based on the spatial distribution of jobs a value between 0 (extrem polycentric) and 1 (extrem monocentric) to define a city's centrality.

\paragraph{Causal Graph Discovery}
To test the hypotheses that urban form affects mobility behavior and to examine the causal relationship between the considered variables in more detail, we apply causal graph discovery. Causal graph discovery is typically based on either constrained- or score-based algorithms. While the former uses conditional independence tests to obtain a graph, the latter defines graph discovery as a structure learning problem, where the graph that maximises some score is selected out of the space of possible graphs. In addition, hybrid approaches that combine both also exist (for an overview and a comparison of different algorithms, refer to \cite{CONSTANTINOU2021,runge2023causal}).

For our problem setting, we adopt a conditional independence-based causal discovery framework, here the PC (Peter-Clark) algorithm \cite{spirtes_glymour_1991,colombo2014order}, which identifies spurious correlations (e.g. resulting from indirect links or common drivers) among the included variables. It determines the presence and absence of edges and their orientation in a DAG using conditional independence tests among all observed variables and VKT. The PC algorithm assumes independent and identically distributed (\textit{iid}) data and can handle linear and nonlinear variable dependencies, depending on the chosen conditional independence test. A detailed introduction to the method can be found in \cite{runge2023causal}. To summarize, the PC algorithm starts from a fully connected undirected graph and consists of three phases: (1) The \emph{skeleton phase} uses conditional independence tests to learn the skeleton of adjacencies of the causal graph. If two variables $X$ and $Y$ are found to be independent conditional on a (possibly empty) set of other variables $\mathbf{Z}$, then the edge between $X$ and $Y$ is removed. The algorithm efficiently selects these conditioning sets to avoid iterating through all possible ones. (2) In the \emph{collider orientation phase} all \emph{collider motifs} of the form $X \rightarrow Y \leftarrow Z$ where $X$ and $Z$ are non-adjacent are detected. These orientations can be inferred because these motifs are only compatible with a particular pattern of (conditional) (in-)dependencies. (3) The final \emph{orientation phase} uses graphical rules based on the assumption of acyclicity to orient as many remaining unoriented edges as possible. Some edges may not be oriented (represented as $o\-o$ in the graph) such that the result of the PC algorithm is often a set of graphs, called the Markov equivalence class, that cannot be distinguished by conditional independence testing alone.

Regarding the choice of a conditional independence test, we assume variable relationships to be continuous and linear with potentially non-gaussian marginal distributions and, hence, select a robust version of the partial correlation independence test where the variables $X,Y,\mathbf{Z}$ are first transformed to have normal marginal distributions before applying the standard partial correlation test with an assumed $t$-null distribution. We also test an alternative conditional independence test that assumes multivariate, continuous variables with more general nonlinear dependencies together with a permutation-based test, but with less detection power for linear dependencies, called CMIknn~\cite{runge_2018_CMIknn}. The results of the five fold validation with CMIknn can be found in SI, Additional results, Fig, \ref{si_fig1_dag_cmiknn_all_seeds}.

Next to discovering a DAG, we use the framework to examine a measure of the strength of an observed causal dependency between an $X$ and $Y$ by investigating the partial correlation value $\rho$ derived from the selected conditional independence tests in the PC algorithm's skeleton phase at the maximal p-value over all considered conditioning sets $\mathbf{Z}$. This partial correlation value between two nodes can be seen as a qualitative estimator of causal strength.

The PC algorithm assumes the Causal Markov Condition, Faithfulness and Causal Sufficiency. The first indicates that a statistical dependence between two variables is the result of an underlying causal connection. Faithfulness states that statistical independence indicates the absence of any causal connection, and causal sufficiency implies that the analyzed variables include all common causes. Here we have to relax the causal sufficiency assumption as not all causes of travel distance can be measured or are freely available as open data. This means that any detected edge could still be rendered as spurious if a relevant common cause was included in the analysis. On the contrary, detecting an absence of a link does not require the causal sufficiency assumption. In addition, we assume the following about our variable relationships: first, we expect that none of the urban form features can be caused by the target VKT. Second, we assume that income differences (as a proxy for socio-demographics and behavioral attitudes) can affect both urban form (via residential self-selection) and travel behavior, but cannot be caused by urban form. Third, as previous studies \cite{naess_2015, SCHEINER2023100820} have frequently mentioned the indirect effect of distance to the city center on other variables, we assume that distance to city center cannot be caused by other variables. These assumptions constrain the PC algorithm's inference and lead to more stable results.

To calculate the DAG, we use a balanced sample of $1542$ TAZ. As some cities have more TAZ than others, the samples are pooled equally from all cities to remove city-specific biases. To improve the robustness of the approach, we randomly sample five rounds with different seeds and only report links that are consistently present across all rounds. When the orientation of a link varies, we adopt the orientation of most appearances (refer to SI, Additional results, Fig, \ref{si_fig1_dag_robustparcorr_all_seeds}). We standardize the features per city by subtracting the mean and scaling to unit variance to remove city-specific hidden confounding, such as differences in gas prices, city sizes or network lengths.

\paragraph{Model Generalisation} 
To analyze how much of the overall variation in VKT and related emissions can be explained by urban form, we adopt a 6-fold, city-wise cross-validation procedure. We train a Gradient Boosting Decision Tree Regression Model (GBDT), using the XGBOOST Python library \cite{xgboost}, on five cities and predict VKT in the unseen 6th city. We choose a GBDT model, as it can represent non-linear dependencies, is robust against feature multicollinearity \cite{ding2018applying}, and has been shown to perform best with data in tabular format in a similar problem setting \cite{spadon2019}. We evaluate the generalization performance using the coefficient of determination (R2), the mean absolute error (MAE), and root mean squared error (RMSE).

\paragraph{Model Interpretation}
To analyse heterogeneity in urban form effects, we calculate causal Shapley values on the prediction test set, as proposed by Heskes et al. (2020) \cite{heskes2020}. Causal Shapley values are based on (marginal) Shapley values \cite{lundberg_shap_2017}, which solve an attribution problem where the prediction score of a model is distributed to its individual features. This is calculated by taking the weighted average across all possible feature combinations, also called coalitions, to evaluate how much a prediction changes when a feature is part of a coalition $S$ versus when it is not $\overline{S}$.  

Estimating the prediction in the absence of a feature sparked a debate among scholars about which probability distribution is the right one to draw from (for further reference, see for example \cite{kaddour2022}), questioning the suitability of Shapley values, especially when correlated features are present. While different solutions were proposed \cite{Frye_2020, Chen_Janizek_Lundberg_2020, yonghan_2022, janzig_2020}, we adopt causal Shapley values which estimate the prediction based on sampling from an interventional conditional distribution instead of the previously used observational conditional distributions.For this, causal Shapley values aim to incorporate the relationships of a causal DAG, however, the current implementations of causal Shapley only consider a causal chain. A causal chain is a sequential representation of a DAG that defines a partial causal ordering, refer to Figure 2 in \cite{heskes2020}). By doing so, causal Shapley values break the dependence between the features in the coalition $S$ and the remaining features, allowing to obtain meaningful feature attributions that reflect the underlying causal structure when explaining a model. Causal Shapley values require a causal chain as an input. In the chain, one feature can only cause subsequent features of the chain and the target, while information about missing edges is inferred from the data based on conditional independencies. Based on the discovered DAG, we adopt the order $distance\ to\ center \to distance\ to\ employment \to population\ density \to street\ connectivity$.

In comparison to the partial correlation value $\rho$ of the causal graph discovery, causal shapley values do not quantify the strength of dependencies between features, but allocate only an importance to the direct effect of a feature on the target, considering inter-feature relationships. For analysing spatially explicit urban form effects, the properties of causal Shapley values are of high use. They allow (1) to keep the desirable property of marginal Shapley values to calculate effects for individual samples (which translate to individual locations) and (2) to reflect causal relationships between our features and the target via causal chain graphs (see SI, Comparison of Causal and Marginal Shapley values, Fig.~\ref{si-fig-2-shap}.

\section*{Acknowledgements and funding sources}
We thank Peter Berril and Ioan Gabriel Bucur for the insightful discussions on the manuscript.
This work received funding from the CircEUlar project of the European Union’s Horizon Europe research and innovation program under grant agreement 101056810.

\section*{Data and code availability}
For all openly available data used in this study references are provided in Methods, Datasets. Upon publication, the code of this work will be provided as a freely accessible git repository here: \url{https://github.com/wagnerfe/xml4uf}.

\newpage
\bibliography{references}

\newpage
\section*{Supplementary Information}

\subsection*{Additional methods material}
\subsubsection*{Location specific emission factors}
Location specific emission factors are approximated based on official statistics about vehicle fleet numbers (share of small-medium and large cars) in the US \cite{statista_us_cars_2021}, Germany \cite{uba_2021} and Latin America \cite{statista_brazil_cars_2023} (note that due to a lack of data availability for Bogotá, we assume Bogotá to have a similar share of car types than Brazil) and respective emissions factors, see summary Table \ref{table_emission_factors}. Shares of vehicle types are derived for US cities from, Berlin from , and Rio and Bogotá from from . Mapping of car categories to small-medium and large car is shown in Table \ref{table_car_types}, whereas well-to-wheel CO$_2$ intensity factors for small-medium and large cars are derived from the International Energy Agency \cite{iea2021}. We acknowledge that this approximates only internal combustion engine car emissions and does not consider for example uptake of electric vehicles in those countries.

\begin{table}[H]
\begin{tabular}{lccccc}
\textbf{Location} & \multicolumn{1}{l}{\textbf{\begin{tabular}[c]{@{}l@{}}Share small-\\ medium cars\end{tabular}}} & \multicolumn{1}{l}{\textbf{\begin{tabular}[c]{@{}l@{}}Emission\\  factor \\ {[}gCO2eq/km{]}\end{tabular}}} & \multicolumn{1}{l}{\textbf{\begin{tabular}[c]{@{}l@{}}Share\\ large cars\end{tabular}}} & \multicolumn{1}{l}{\textbf{\begin{tabular}[c]{@{}l@{}}Emission\\ factor \\ {[}gCO2eq/km{]}\end{tabular}}} & \multicolumn{1}{l}{\textbf{\begin{tabular}[c]{@{}l@{}}Combined Emission\\ factor \\ {[}gCO2eq/km{]}\end{tabular}}} \\ \hline
Germany           & 0,74                                                                                            & 148                                                                                                        & 0,26                                                                                    & 211                                                                                                       & 164,33                                                                                                             \\
United States     & 0,19                                                                                            & 148                                                                                                        & 0,81                                                                                    & 211                                                                                                       & 199,07                                                                                                             \\
Latin America     & 0,64                                                                                            & 148                                                                                                        & 0,36                                                                                    & 211                                                                                                       & 170,68                                                                                                            
\end{tabular}
\caption{\textbf{Location specific emission factors per kilometer}}
\label{table_emission_factors}
\end{table}

\begin{table}[H]
\begin{adjustbox}{width=0.8\columnwidth,center}
\begin{tabular}{ll|cl|ll}
\multicolumn{2}{c|}{\textbf{Latin America}} & \multicolumn{2}{c|}{\textbf{Germany}}                           & \multicolumn{2}{c}{\textbf{US}}       \\
\textbf{Categorie}    & \textbf{Mapping}    & \multicolumn{1}{l}{\textbf{Categorie}} & \textbf{Mapping}       & \textbf{Categorie} & \textbf{Mapping} \\ \hline
Subcompact            & small               & Mini                                   & Small                  & Crossover          & large            \\
Mid-size              & medium              & Kleinwagen                             & Small                  & Pickup             & large            \\
Compact               & medium              & Kompaktklasse                          & lower-medium           & Small car          & small            \\
Full size             & large               & Mittelklasse                           & lower-medium           & SUV                & large            \\
SUV                   & large               & Obere Mittelklasse                     & lower-medium           & Midsized car       & medium           \\
Compact SUV           & large               & Oberklasse                             & large                  & Luxury car         & large            \\
Microcar              & small               & SUV                                    & large                  & Van                & large            \\
Pickup                & large               & VAN                                    & large                  & Large Car          & large            \\
Luxury car            & large               & Utilities                              & \multicolumn{1}{c|}{-} &                    &                  \\
light commercial      & -                   & Sonstige                               & \multicolumn{1}{c|}{-} &                    &                  \\
Sport car             & large               & \multicolumn{1}{l}{}                   &                        &                    &                  \\
campervan             & large               & \multicolumn{1}{l}{}                   &                        &                    &                  \\
minivan               & medium              & \multicolumn{1}{l}{}                   &                        &                    &                  \\
large van             & large               & \multicolumn{1}{l}{}                   &                        &                    &                  \\
other                 & -                   & \multicolumn{1}{l}{}                   &                        &                    &                  \\
not defined           & -                   & \multicolumn{1}{l}{}                   &                        &                    &                 
\end{tabular}
\end{adjustbox}
\caption{\textbf{Mapping of country specific car types}}
\label{table_car_types}
\end{table}

\newpage

\subsubsection*{Feature distribution}
Feature distributions across all cities are summarised in Table \ref{table_feature_distribution}. 

\begin{table}[H]
\centering
\begin{tabular}{llrrrr}
\toprule
City & Feature & Mean & $\sigma$ & q25 & q75 \\
\midrule
Berlin & Distance to Center & 14166.5 & 9257.0 & 7093.0 & 18376.4 \\
 & Distance to Employment & 5149.0 & 5672.8 & 1201.3 & 7052.8 \\
 & Population Density & 6339.1 & 5726.0 & 1251.2 & 9619.7 \\
 & Street Connectivity & 37.2 & 24.1 & 19.4 & 50.9 \\
Boston & Distance to Center & 15544.0 & 11206.2 & 7149.0 & 21457.4 \\
 & Distance to Employment & 6607.9 & 6975.2 & 1909.0 & 9384.2 \\
 & Population Density & 4255.7 & 4741.0 & 914.7 & 6071.7 \\
 & Street Connectivity & 69.6 & 46.2 & 31.5 & 102.1 \\
Los Angeles & Distance to Center & 23761.7 & 11785.2 & 14769.8 & 32594.3 \\
 & Distance to Employment & 9403.5 & 6450.0 & 4438.5 & 13201.5 \\
 & Population Density & 4923.3 & 3918.4 & 2481.0 & 6372.8 \\
 & Street Connectivity & 44.7 & 17.7 & 33.9 & 55.5 \\
Bay Area & Distance to Center & 41713.4 & 28089.8 & 18229.6 & 67897.9 \\
 & Distance to Employment & 7311.2 & 5608.9 & 3334.6 & 9715.7 \\
 & Population Density & 4433.1 & 4579.8 & 1965.1 & 5372.2 \\
 & Street Connectivity & 56.0 & 26.6 & 40.5 & 68.5 \\
Rio de Janeiro & Distance to Center & 26148.5 & 16562.4 & 12763.6 & 34840.1 \\
 & Distance to Employment & 7269.3 & 6696.3 & 2867.7 & 9329.7 \\
 & Population Density & 8518.4 & 6697.4 & 3091.5 & 12248.7 \\
& Street Connectivity & 63.1 & 39.3 & 34.2 & 85.5 \\
Bogotá & Distance to Center & 13827.1 & 6536.4 & 7917.3 & 18521.2 \\
 & Distance to Employment & 3977.9 & 2510.9 & 1940.0 & 5660.9 \\
 & Population Density & 12248.2 & 2982.3 & 11179.4 & 14259.5 \\
 & Street Connectivity & 159.9 & 96.5 & 91.2 & 213.3 \\
\bottomrule
\end{tabular}
\caption{\textbf{Feature distribution per city.} Provided are the feature average, the standard deviation ($\sigma$), the 25th (q25) and 75th (q75) quantile.}
\label{table_feature_distribution}
\end{table}

\subsection*{Causal Graph Discovery}

\subsubsection*{Urban form effect in previous work}
Urban form effects in previous work. The literature review was conducted via google scholar, arxivx and semantic scholar with a focus on quantitative or mixed methods approaches, urban car travel and the 6D framing. The resulting list is not exhaustive and aims to summarize the most relevant previous work. Across the examined studies we find distance to center and to jobs having positive correlations across all studies that included the feature (note that some studies calculated distance to subcenter or job accessibility instead of distance to jobs. For job accessibility a negative correlation with VKT aligns with a positive correlation of distance to jobs with VKT, as high access implies short distances). Population density was found to have negative correlations and in one study no clear effect could be found. Across all studies that included street connectivity, either measured as intersection density, network connectivity, road density or number of 4 way intersection count, negative coreelations with VKT were found. For income, in 6 cases positive and in 2 cases negative correlations were found. We interpret this as an overall positive correlation with VKT in Fig.~1 of the main manuscript.
Additionally, some studies \cite{SCHEINER2023100820, naess_2015, naess2021} have pointed out that distance to center indirectly affects all other urban form features, reflected as an indirect link Fig.~1 of the main manuscript. We also note that one study pointed out a potential indirect link between density and distance to subcenters/job centers, see Fig.~1 in \cite{naess2021}.

\begin{table}[H]
    \begin{adjustbox}{width=\columnwidth,center}
    \centering
    \begin{tabular}{|l|l|l|l|l|l|}
    \hline
        \textbf{Location} & \textbf{Destination Accessibility} & \textbf{Density} & \textbf{Design} & \textbf{Demographics} & \textbf{Reference} \\ \hline
        Oslo & Distance to CBD (+) & Population Density (-) & ~ & Income (+) & \protect\cite{ding2018applying} \\ 
        ~ & Distance to sub-center (+) & Job Density (-) & ~ & ~ & ~ \\ \hline
        Oslo & Distance to CBD (+) & Population Density (-) & ~ & Income (+) & \protect\cite{naess2017d} \\ 
        ~ & Distance to local CBDs (+) & Job Density (-) & ~ & ~ & ~ \\ 
        Stavanger & Distance to CBD (+) & Jobs/Capita Density (+) & ~ & Income (+) & ~ \\ \hline
        Montreal & ~ & Population Density (-) & ~ & Income (+) & \protect\cite{ZAHABI20151} \\ \hline
        US cities & Distance to CBD (+) & Job Density (-) & Network Connectivy (-) & ~ & \protect\cite{nasri2012} \\ 
        ~ & ~ & Population Density (-) & ~ & ~ & ~ \\ \hline
        US cities & Distance to CBD (+) & Job Density (x) & Intersection Density (-) & ~ & \protect\cite{ewing2017does} \\ 
        ~ & Job Accessibility (-) & Population Density (-) & 4-way Intersection (-) & ~ & ~ \\ \hline
        US cities & Distance to CBD (+) & Job Density (x) & Intersection Density (-) & ~ & \protect\cite{stevens2017does} \\ 
        ~ & Job Accessibility (-) & Population Density (-) & 4-way Intersection (-) & ~ & ~ \\ \hline
        Berlin & Distance to CBD (+) & Population Density (x) & ~ & Income (-) & ~ \\ \hline
        ~ & Distance to sub-center (+) & ~ & ~ & ~ & \protect\cite{WAGNER2022} \\ \hline
        Washington & Distance to CBD (+) & ~ & Street Conectivity (-) & Income (+) & \protect\cite{dingchuan2014} \\ \hline
        US cities & Distance to Jobs (+) & Population Density (-) & ~ & Income (+) & \protect\cite{ewingtian2015} \\ \hline
        Germany & Distance to Jobs (+) & Commercial Density (-) & Road Density (-) & Income (-) & \protect\cite{Vance2007} \\ \hline
        Chicago & Distance to CBD (+) & Population Density (-) & ~ & ~ & \protect\cite{LINDSEY2011} \\ \hline
        Switzerland & ~ & Population Density (-) & ~ & ~ & \protect\cite{THAO2020} \\ 
        ~ & ~ & Job Density (-) & ~ & ~ & ~ \\ \hline
    \end{tabular}
    \end{adjustbox}
\caption{\textbf{Effects of Destination Accessibility, Density, and Demographics in previous work} For every study it is marked whether the feature was (among others) included and a positive observed effect on VKT is marked with (+), a negative with (-) and no clear effect found with (x). }
\label{table_feat_source_list}
\end{table}

\newpage
\subsubsection*{Causal DAG based on robust partial correlation conditional independence test across all seeds}
\begin{figure}[H]
\centering
\includegraphics[width=0.8\textwidth]{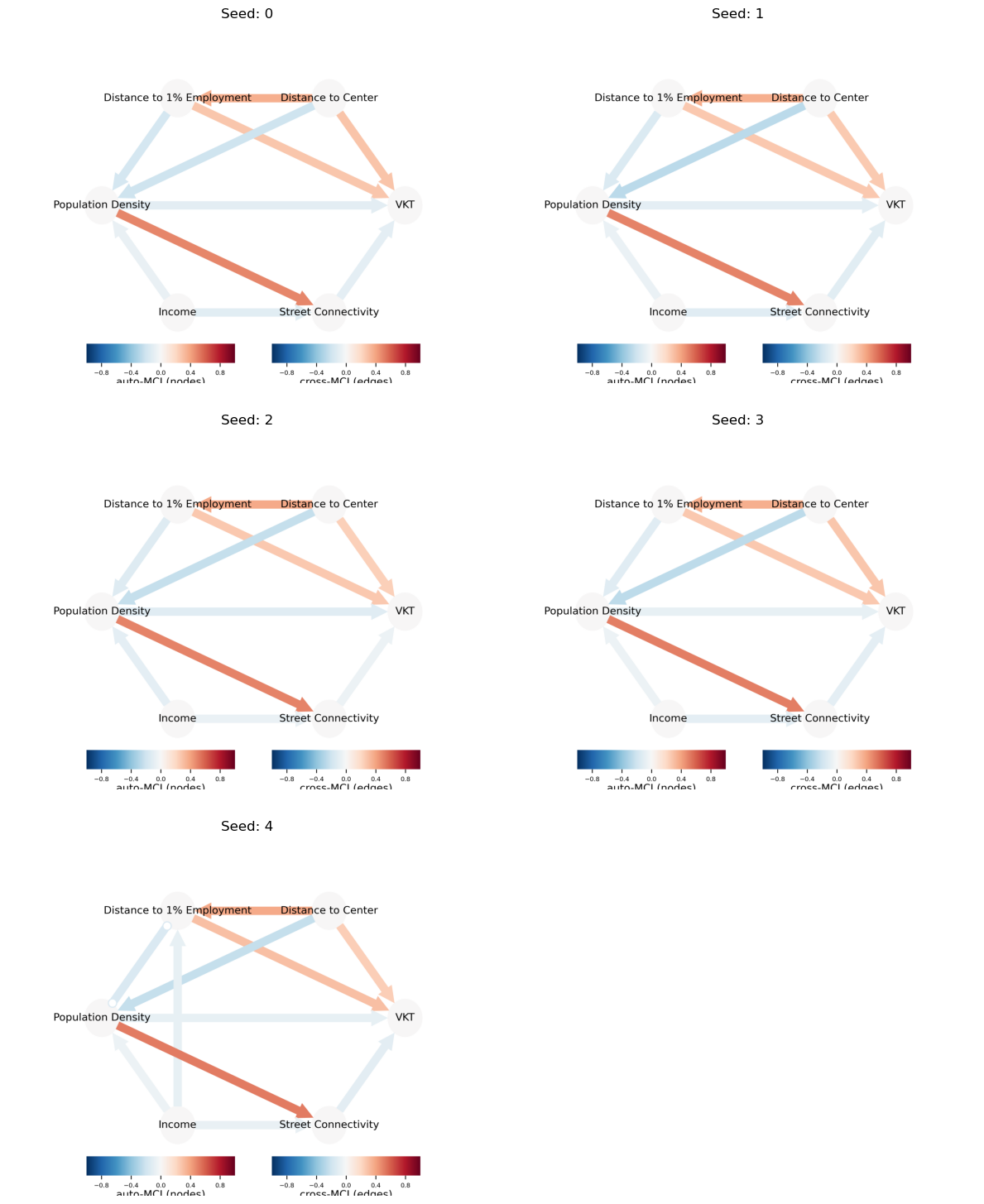}
\caption{\textbf{Causal DAG based on robust partial correlation conditional independence test across five seeds.  Edges whose direction cannot oriented are represented as $o\-o$ in the graph.
}}
\label{si_fig1_dag_robustparcorr_all_seeds}
\end{figure}

As we create a balanced sample of $1542$ TAZ, representing each city equally, we can calculate a DAG five times with a random sample of different seeds as shown in Fig.~\ref{si_fig1_dag_robustparcorr_all_seeds} and create a summary graph, considering only consistently present links. This implies that we do not report the link from income to distance to employment in seed 4. Additionally, we report the link from distance to employment to population density, as it is present in all sampling rounds, but the direction is not solved only one time, see seed 4.

\subsubsection*{Causal DAG based on cmiknn conditional independence test across all seeds}

\begin{figure}[H]
\centering
\includegraphics[width=0.8\textwidth]{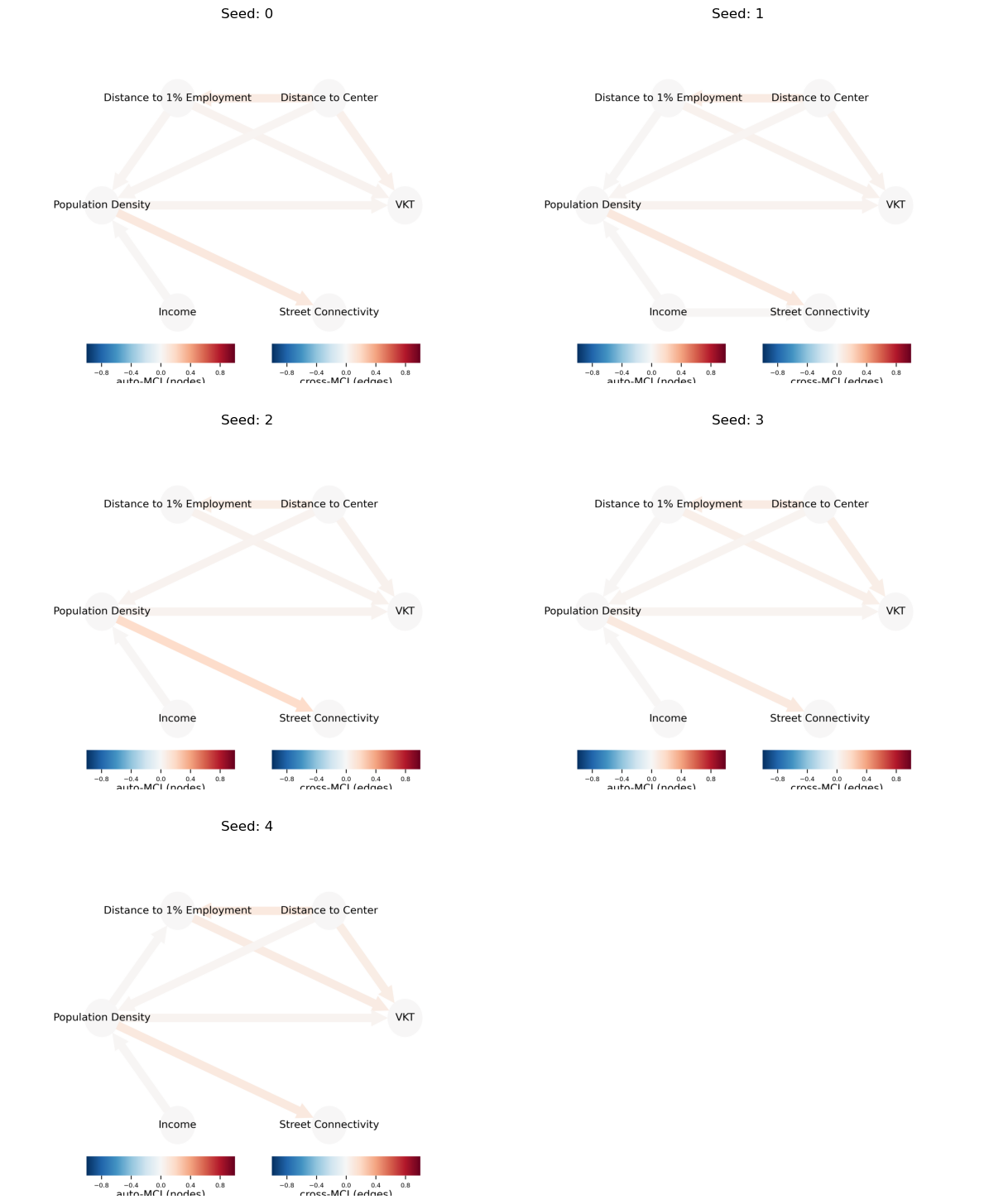}
\caption{\textbf{Causal DAG based on cmiknn conditional independence test across five seeds.
}}
When using cmiknn conditional independece test, we also measure a direct effect of distance to center, distance to employment and density on VKT, but we do not find a significant link between street connectivity and VKT. Differences in indirect feature links are observed in seed 2 and seed 4 between distance to employment and density as well as between income and street connectivity which is only present in one out of 5 sampling rounds.

\label{si_fig1_dag_cmiknn_all_seeds}
\end{figure}

\newpage
\subsection*{Discussion of causal graph assumptions}
One important assumption (which has also been present in previous studies), is the choice and definition of included 6Ds. The feature choice was constrained by data availability and modelling dimensionality. Our features are proxies for measurable and non-measurable urban form and demographic differences in cities. As dimensionality affects link detection power \cite{runge_2019_science}, we aimed at finding a minimal set of representative features yet recognize that this comes at the cost of not accurately describing urban form. Further, the 6Ds have been defined in various ways in the past. Here, we chose definitions that, to our knowledge, represent the true effect of one feature in a most meaningful way. For example, as the zip code sizes vary within a city, we scaled all density features per area of zip code. Yet, this scaling can affect the resulting DAG and requires interpreting all remaining results only on the basis of our definitions.

Another key assumption is the choice of conditional independence (CI) test. We acknowledge that we observe both linear and non-linear dependencies between the analyzed variables (see SI Fig.~\ref{si/data_dependencies}). This implies that using a CI test that supports both linear and non-linear dependencies (e.g. GPDC , CMIknn \cite{runge_2019_science,runge_2018_CMIknn}) might be best suited for the analysis. Yet, they come at the cost of having a lower detection power for linear links, and hence true linear, but smaller effects might be overlooked. We acknowledge that this is a current shortcoming of the method and hence, provide the DAG of Fig.~\ref{fig1_dag} also calculated with CMIknn CI test, displayed in SI Fig.~\ref{si_fig1_dag_cmiknn_all_seeds}. Future work can compare different graph discovery algorithms to get a more robust estimate of the true DAG.

\subsection*{Model Interpretation}
\subsubsection*{Model generalization}
We find that all 6Ds with a direct causal effect on VKT hold predictive information on car commute distances when generalizing towards an unseen city. When training a model in five cities on all features with a direct causal effect on the target and predicting in an unseen 6th city, we can show that the variation in VKT can be explained up to 84\% in Berlin, 62\% in Boston, 51\% in Bogotá, 21\% in Los Angeles, 41\% in Rio de Janeiro, and 26\% in the Bay Area, as shown in Table \ref{table_ml_results}. 

\begin{table}[!ht]
    \centering
    \begin{tabular}{lllllll}
        \textbf{city} & \textbf{R2 Train} & \textbf{R2 Test} & \textbf{MAE [km]} & \textbf{RMSE [km]} & $\overline{\textbf{VKT}}$ \textbf{[km]} & \textbf{ $\sigma$  [km]} \\ 
        ~ & 5 cities & 1 city & ~ & ~ & ~ & ~ \\ \hline 
        Berlin & 0.42 & 0.84 & 1.81 & 2.52 & 12.53 & 9.18 \\ 
        Boston & 0.41 & 0.62 & 1.22 & 1.59 & 7.56 & 3.25 \\ 
        Los Angeles & 0.55 & 0.21 & 2.22 & 3.33 & 15.24 & 3.74 \\ 
        Rio de Janeiro & 0.42 & 0.41 & 2.59 & 3.73 & 12.38 & 4.86 \\ 
        Bay Area & 0.47 & 0.26 & 1.90 & 2.55 & 14.35 & 2.97 \\ 
        Bogotá & 0.41 & 0.51 & 1.75 & 2.11 & 11.84 & 3.03 \\ 
    \end{tabular}
\caption{\textbf{Results of a 6-fold, city-wise cross-validation procedure to analyze how much of the variation in VKT can be explained by urban form.} The generalization performance is provided for each city representing the test set, using the R2 score, the mean average error (MAE) and the root mean square error (RMSE). Additionally, the average VKT of the city is provided for reference.}
\label{table_ml_results}
\end{table}

When training in five cities and predicting in a 6th city (for more details see Methods, Model Generalization), the Mean Average Error Ranges (MAE) from $1.75$ km in Bogotá to $2.59$ km in Rio, while the Root Mean Square Error (RMSE) varies between $1.59$ km in Boston and $3.73$ km in Rio de Janeiro. While the high R2 score in Berlin is also due to the very high standard deviation of VKT, it also shows that given our model and data, the variation in VKT can be better explained by the urban form in more mono-centric cities, such as Berlin or Boston.

\newpage
\subsubsection*{Comparison of Causal and Marginal Shapley Values}
\begin{figure}[H]
\centering
\includegraphics[width=0.7\textwidth]{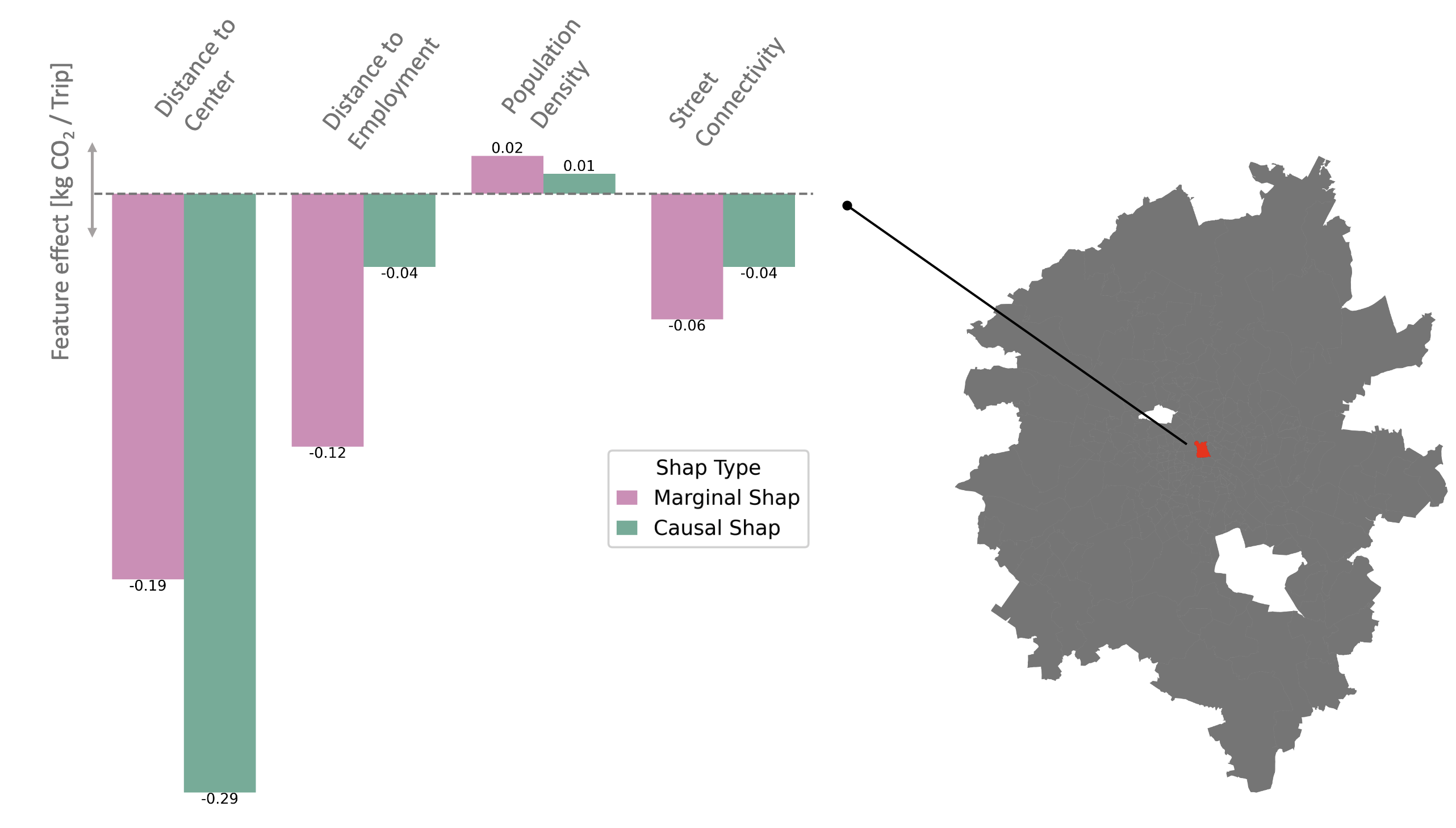}
\caption{\textbf{Comparison of causal and marginal Shapley values} Comparison of the differences in feature importance for traditional marginal (pink) and causal (green) causal Shapley values for an example location in Berlin.} 
\label{si-fig-2-shap}
\end{figure}

The comparison of marginal and causal Shapley values for an example location in Berlin is shown in Fig.~\ref{si-fig-2-shap}. It highlights how causal shap (green bars) allocates most importance to distance to center and removes importance from other features like employment, density or street connectivity for this location in comparison to marginal shapley values (pink bars).

\subsubsection*{Relative feature importance per city}

\begin{figure}[H]
\centering
\includegraphics[width=0.7\textwidth]{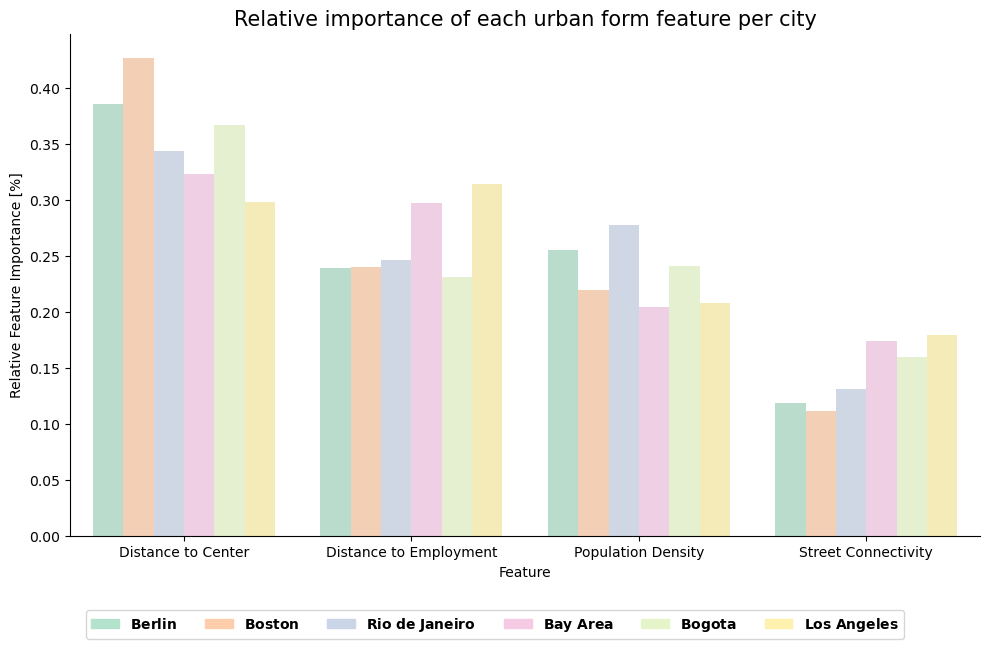}
\caption{\textbf{Relative feature importance per city} For each city we calculate the relative feature importance based on the mean absolute causal shapley values per feature.} 
\label{si-fig2-bar}
\end{figure}

\newpage
\subsubsection*{Destinations of areas with high emissions due to long distances to the center}
\begin{figure}[H]
\centering
\includegraphics[width=1\textwidth]{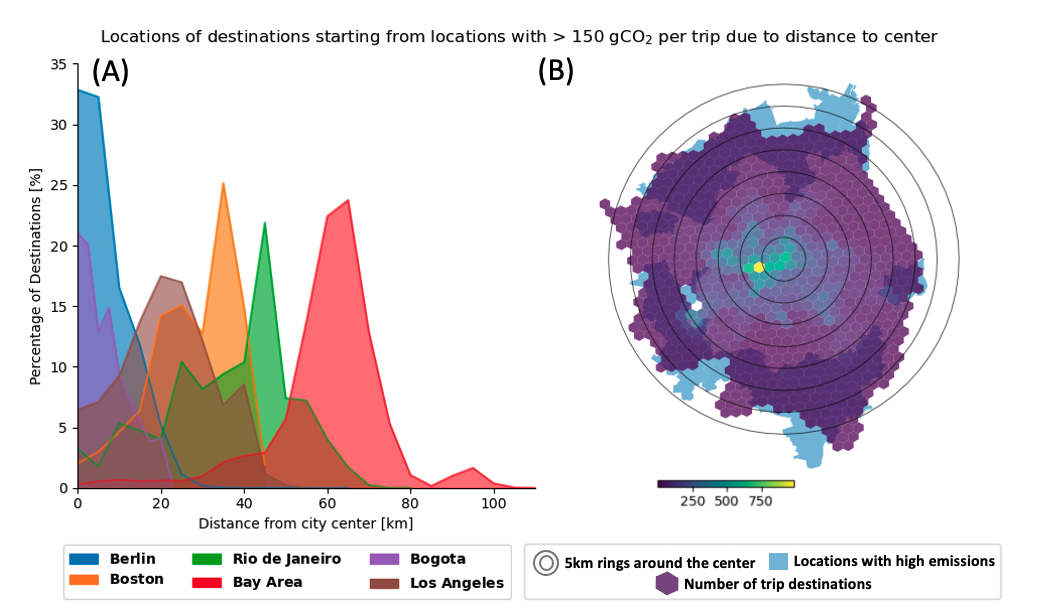}
\caption{\textbf{Destinations of locations with high emissions due to their location relative to the city center} Panel A: Percentage of destinations in 5km rings around the city center. Panel B: Exemplary visualization of 5km rings (black), locations with >150 gCO$_2$ emissions (blue) and trip destinations in hexagons, where yellow indicates many destinations and purple few.} 
\label{si_fig2_A}
\end{figure}

To further assess why the effect of distance to center follows different magnitudes, we analyze where trips go to from all locations with feature effects (causal Shapley values) of above $150$ gCO$_2$ due to their location relative to the center. To do so, we define $5$ km rings around the city center and calculate the percentage of how many trips starting in locations in the very outskirts end in each ring, as shown in Fig.~\ref{si_fig2_A}. We find that in Berlin and Bogotá, most trips also from remote locations travel to locations close to the city center (compare Fig.~\ref{si_fig2_A} descending slope with higher distance to the center). Yet, in other cities, especially in the Bay Area, Boston and Rio, there are other destinations than the center that attract many trips. In the Bay Area (red) the peak at $60$ to $65$ km marks San Jose, while in Boston, we find many destinations in Gloucester at $35$km distance from the center and in Rio de Janeiro we find many destinations in Santa Cruz in the west of the municipal boundaries at $45$ km to the center.

\newpage
\subsubsection*{Areas with very high population densities in Los Angeles}
\begin{figure}[H]
\centering
\includegraphics[width=0.6\textwidth]{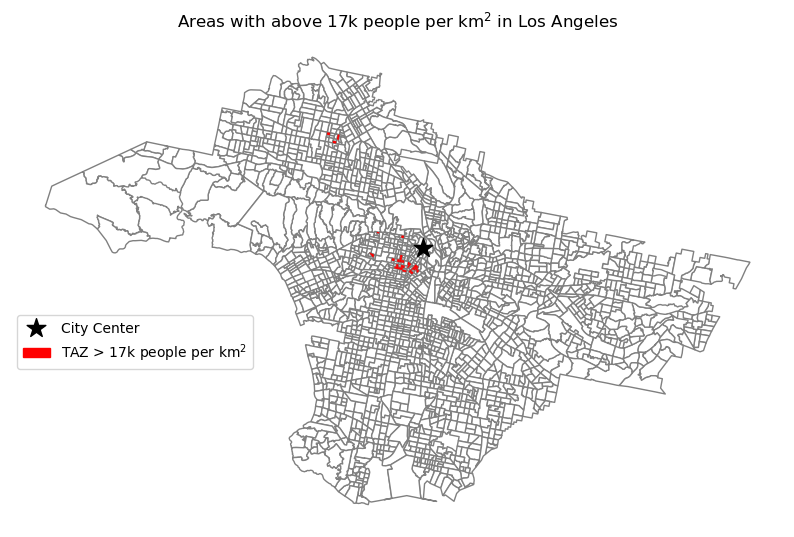}
\caption{\textbf{Locations with very high population densities in Los Angeles} Areas with high population densities marked in red, while the city center is marked with a black star.} 
\label{si_fig2_C}
\end{figure}

To better understand at which locations in Los Angeles causal Shapley values turn positive for high population densities, we plot all locations with population density values over 17k people per $km^2$ in Fig.~\ref{si_fig2_C}. We find that most locations are located very close to downtown, in the neighbourhoods Koreatown and Westlake, as well as some in the neighborhood Panorama City.

\subsubsection*{Areas with very low street connectivity values in Los Angeles}
\begin{figure}[H]
\centering
\includegraphics[width=0.6\textwidth]{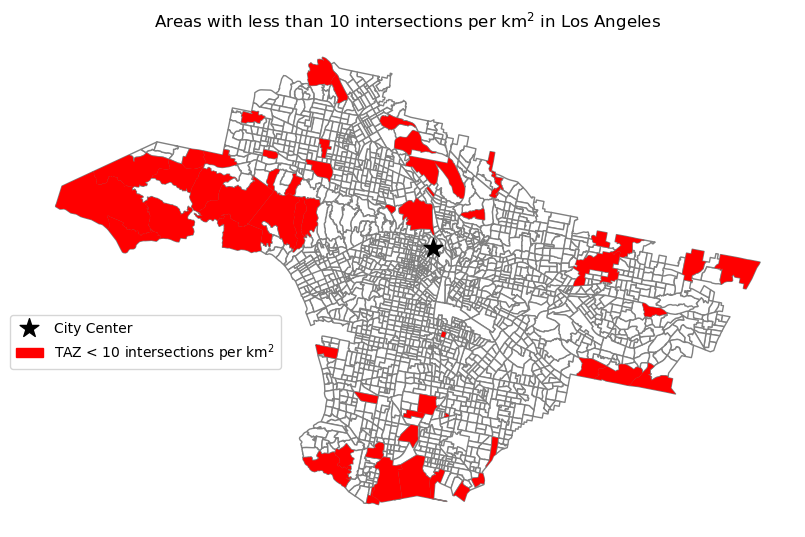}
\caption{\textbf{Locations with very low street connectivity in Los Angeles} Areas with low street connetivity values are marked in red, while the city center is marked with a black star.} 
\label{si_fig2_D}
\end{figure}

To better understand at which locations in Los Angeles we receive very high causal Shapley values for low street connectivities, we plot all locations with street connectivity values below 10 intersections per $km^2$ in Fig.~\ref{si_fig2_D}. We find that most locations are located at the outskirts, such as in Topanga State Park, Malibu Creek State Park, Chino Hills State Park but also at the port in Long Beach.

\newpage
\subsubsection*{Individual feature effects per city}
The following displays the causal Shapley effects, summarized in Fig.~\ref{fig2} as individual plots per city.
\begin{figure}[H]
\centering
\includegraphics[width=0.7\textwidth]{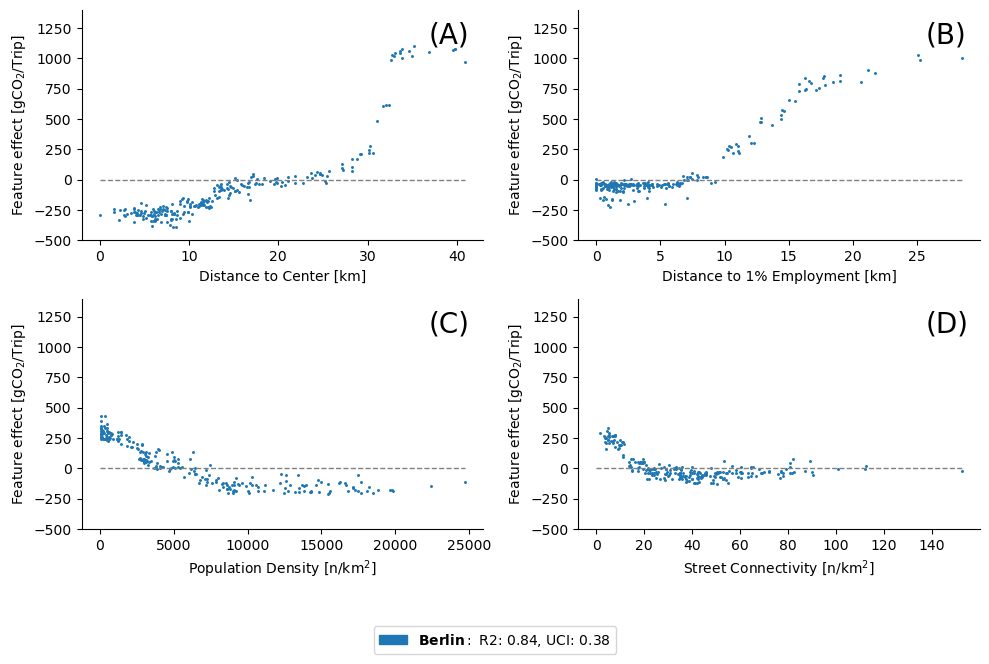}
\caption{\textbf{Causal Shapley effects over feature distributions in Berlin}} 
\label{si_fig2-ber}
\end{figure}

\begin{figure}[H]
\centering
\includegraphics[width=0.7\textwidth]{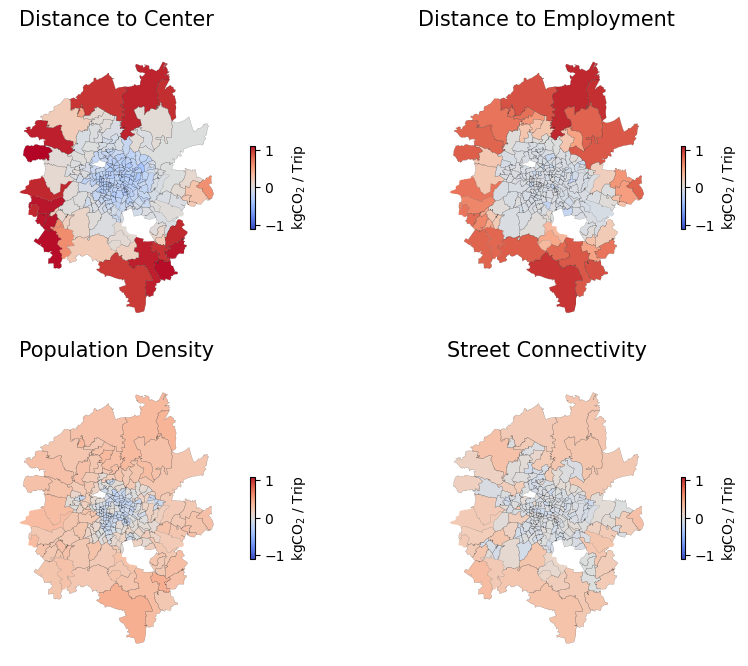}
\caption{\textbf{Causal Shapley effects spatially in Berlin}} 
\label{si_fig2-ber-map}
\end{figure}

\newpage
\begin{figure}[H]
\centering
\includegraphics[width=0.7\textwidth]{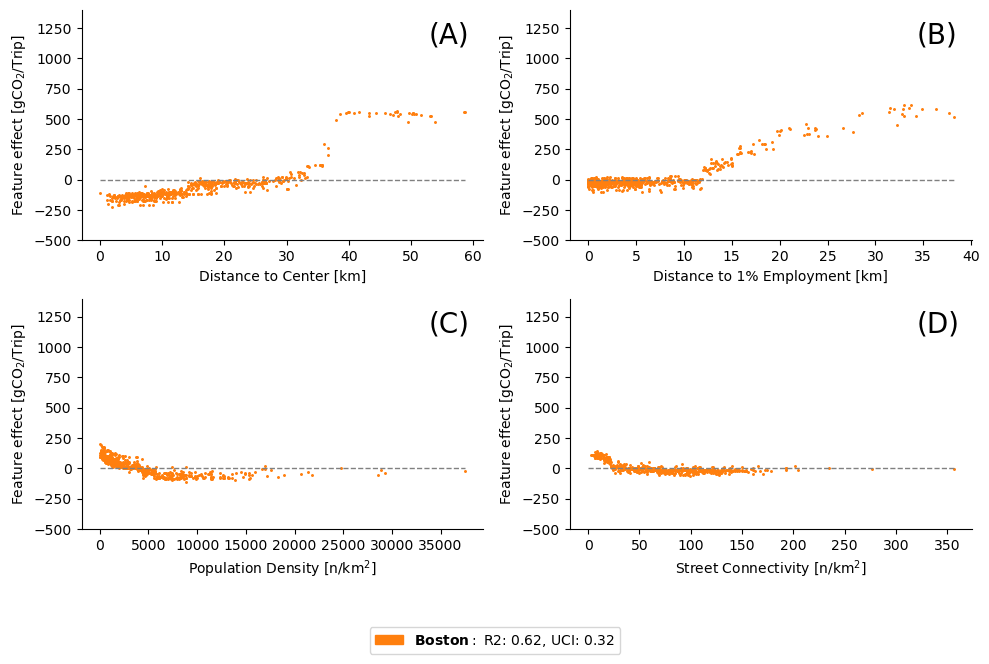}
\caption{\textbf{Feature effects in Boston}} 
\label{si_fig2-bos}
\end{figure}

\begin{figure}[H]
\centering
\includegraphics[width=0.7\textwidth]{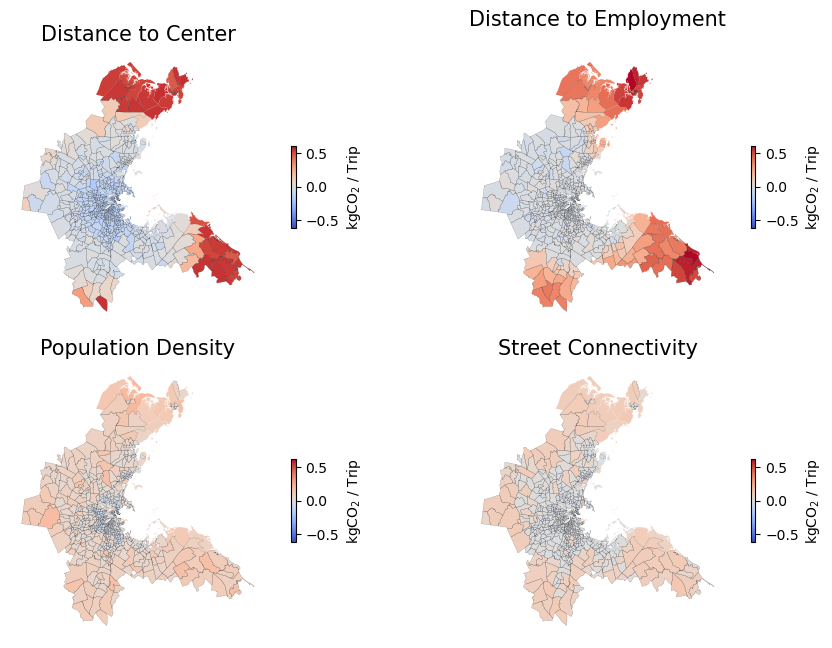}
\caption{\textbf{Causal Shapley effects spatially in Boston}} 
\label{si_fig2-bos-map}
\end{figure}

\newpage
\begin{figure}[H]
\centering
\includegraphics[width=0.7\textwidth]{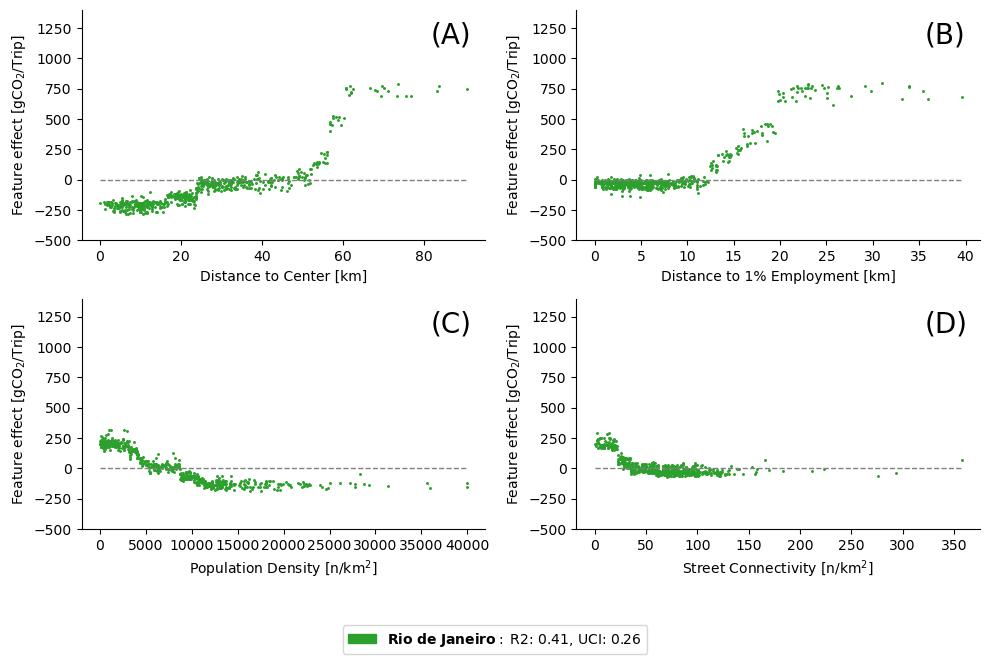}
\caption{\textbf{Feature effects in Rio de Janeiro}} 
\label{si_fig2-rio}
\end{figure}

\begin{figure}[H]
\centering
\includegraphics[width=1\textwidth]{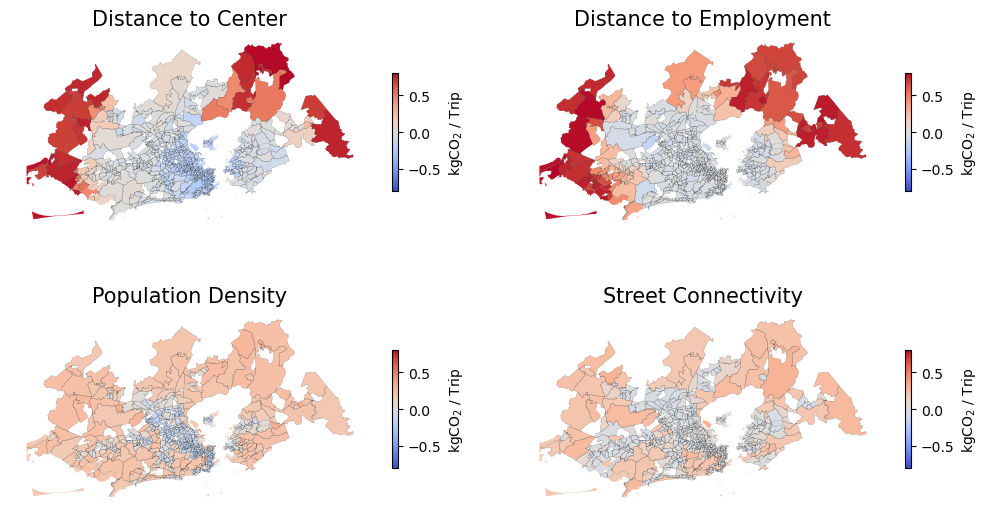}
\caption{\textbf{Causal Shapley effects spatially in Ro de Janeiro}} 
\label{si_fig2-rio-map}
\end{figure}

\newpage
\begin{figure}[H]
\centering
\includegraphics[width=0.7\textwidth]{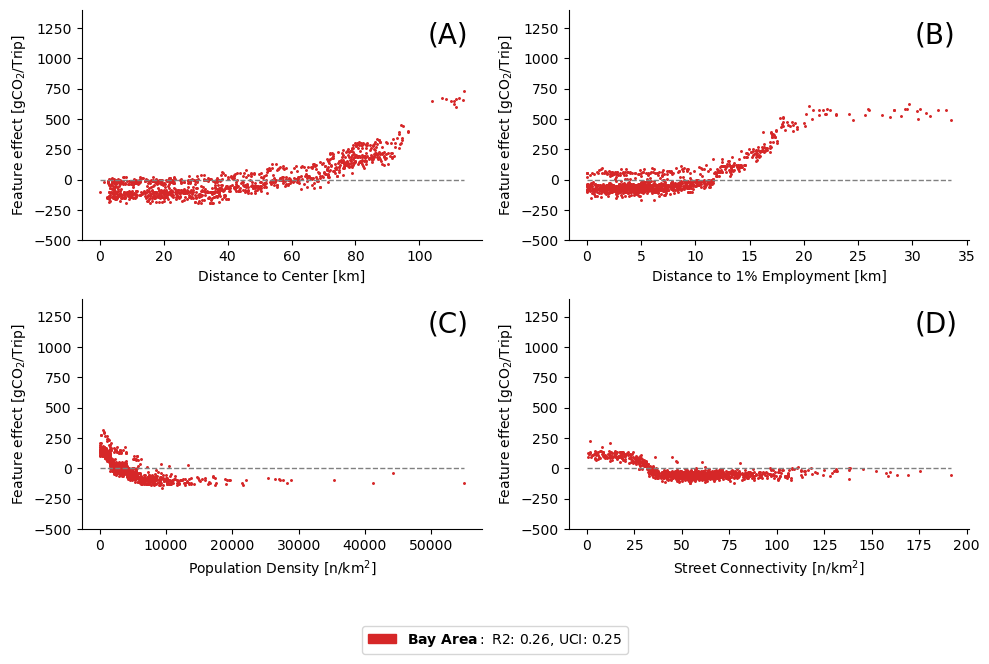}
\caption{\textbf{Feature effects in the Bay Area}} 
\label{si_fig2-sfo}
\end{figure}

\begin{figure}[H]
\centering
\includegraphics[width=0.7\textwidth]{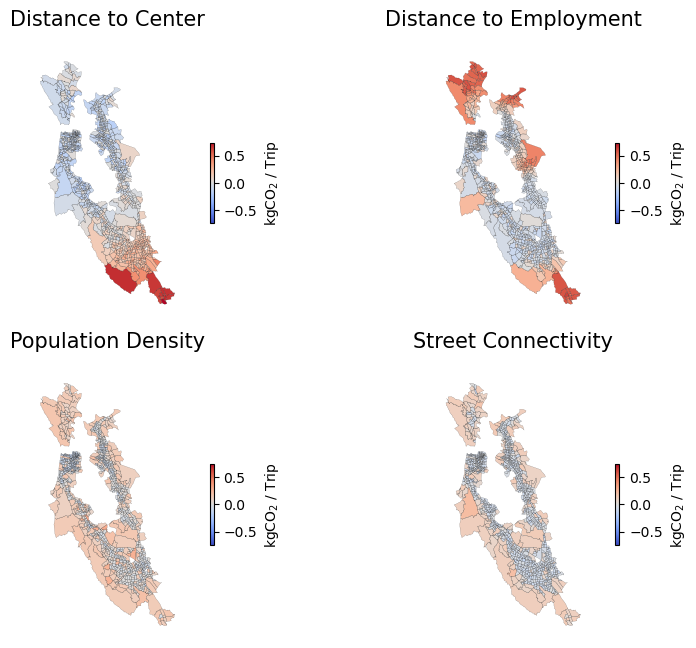}
\caption{\textbf{Causal Shapley effects spatially in the Bay Area}} 
\label{si_fig2-sfo-map}
\end{figure}

\newpage
\begin{figure}[H]
\centering
\includegraphics[width=0.7\textwidth]{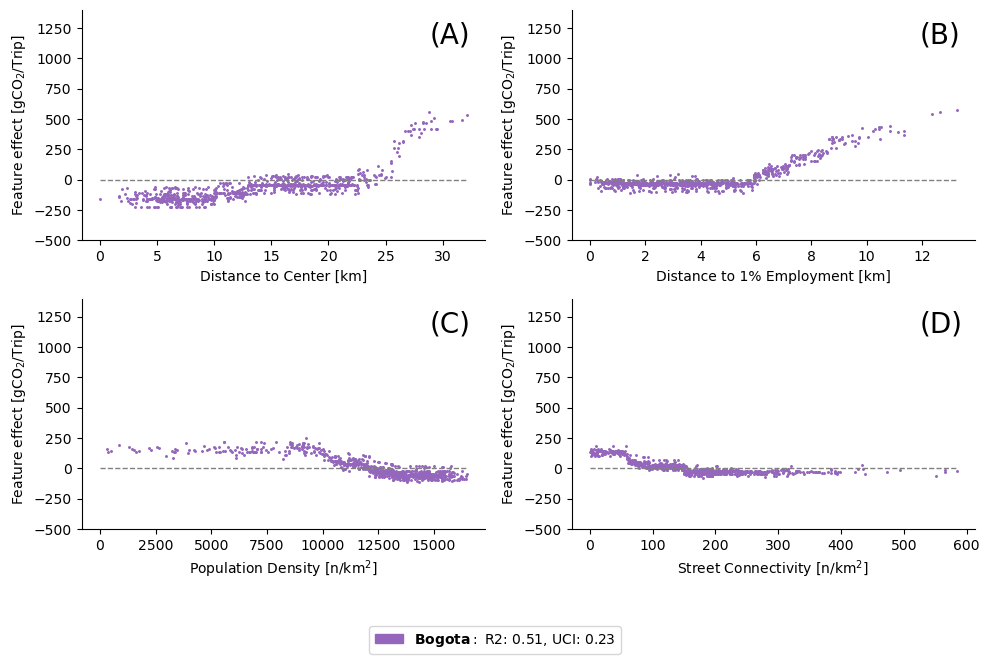}
\caption{\textbf{Feature effects in Bogotà}} 
\label{si_fig2-bog}
\end{figure}

\begin{figure}[H]
\centering
\includegraphics[width=0.7\textwidth]{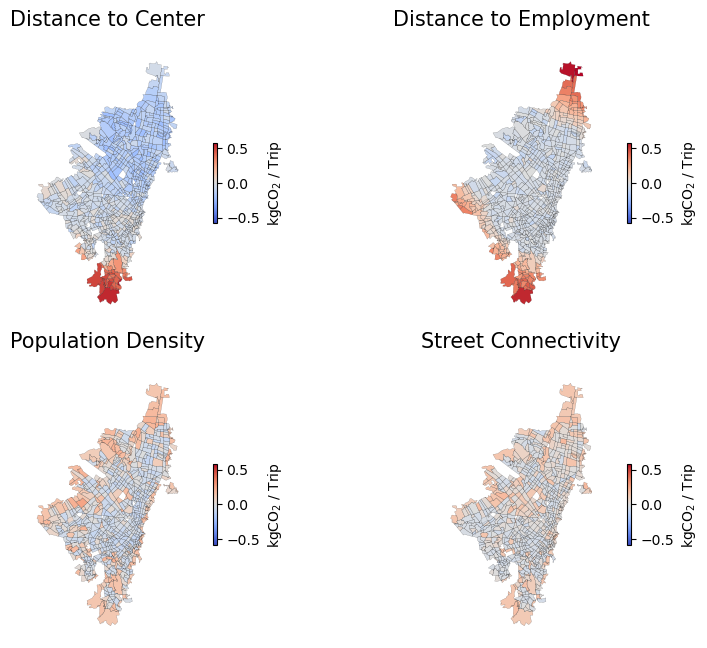}
\caption{\textbf{Causal Shapley effects spatially in Bogotá}} 
\label{si_fig2-bog-map}
\end{figure}

\newpage
\begin{figure}[H]
\centering
\includegraphics[width=0.7\textwidth]{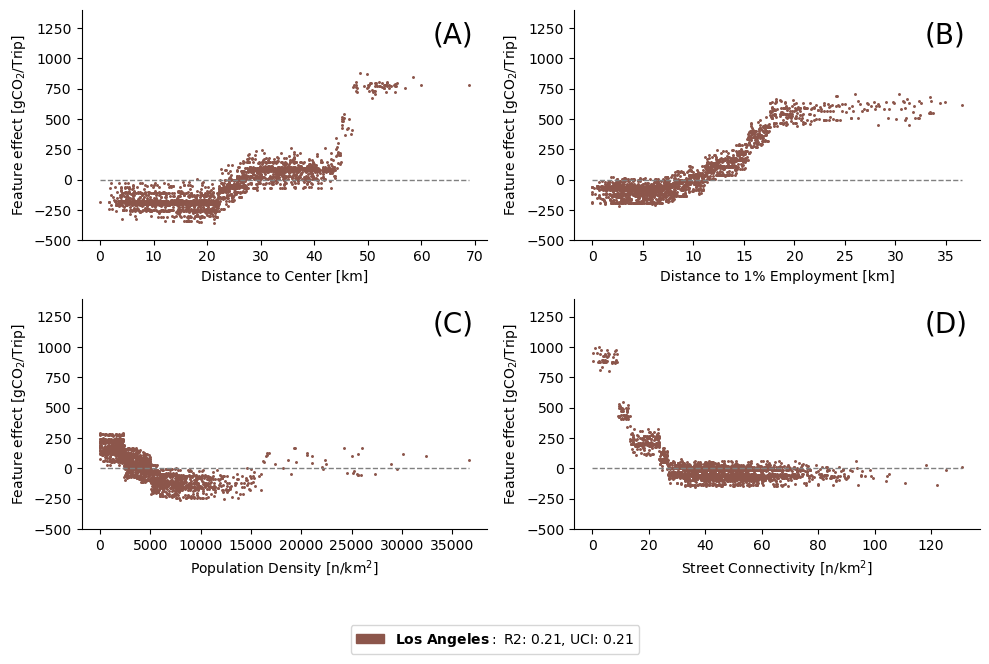}
\caption{\textbf{Feature effects in Los Angeles}} 
\label{si_fig2-bog}
\end{figure}

\begin{figure}[H]
\centering
\includegraphics[width=1\textwidth]{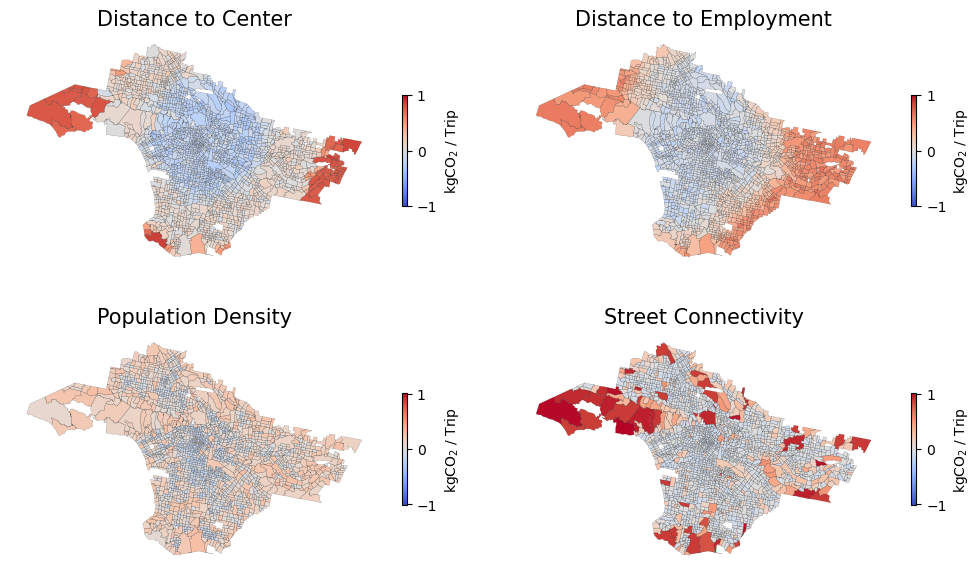}
\caption{\textbf{Causal Shapley effects spatially in Los Angeles}} 
\label{si_fig2-lax-map}
\end{figure}

\newpage
\subsubsection*{Individual feature effects in the Bay Area with 4 centers}
The following displays the causal Shapley effects, when considering 4 centers in the Bay Area, located in Downtown San Francisco, San Jose, Fremont and Oakland.
\begin{figure}[H]
\centering
\includegraphics[width=0.7\textwidth]{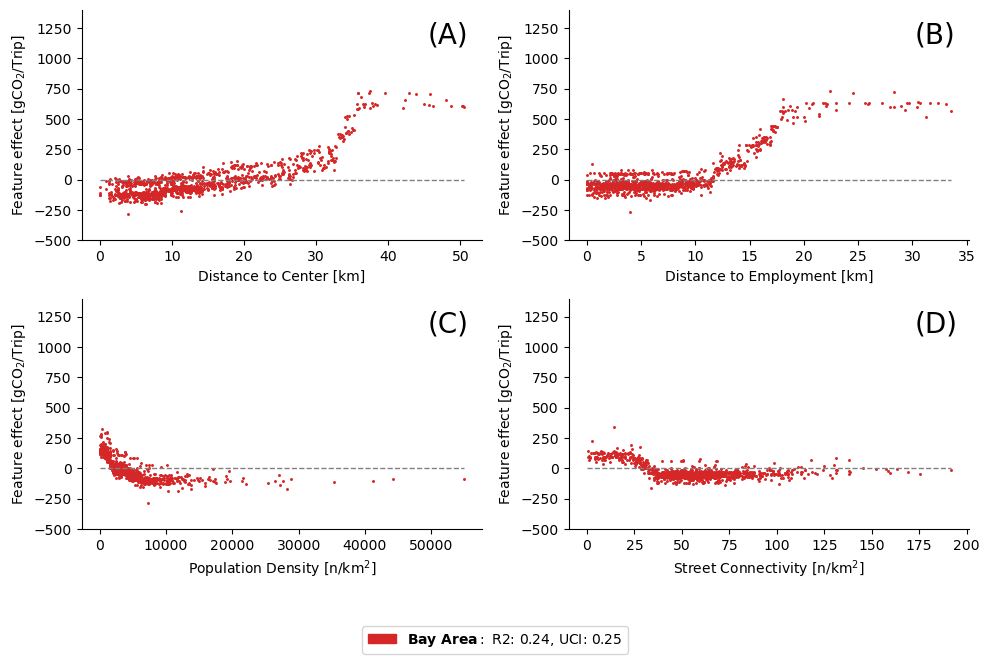}
\caption{\textbf{Causal Shapley effects over feature distributions in Berlin}} 
\label{si_fig2-sfo_4cbd}
\end{figure}

\newpage
\subsubsection*{Threshold effects of distance to center and population density}
Spatially explicit differences in urban form effects of density and accessibility, displayed in remaining four cities, A) Boston, B) Bay Area, C) Bogotá and D) Los Angeles. 

\begin{figure}[H]
\centering
\includegraphics[width=0.7\textwidth]{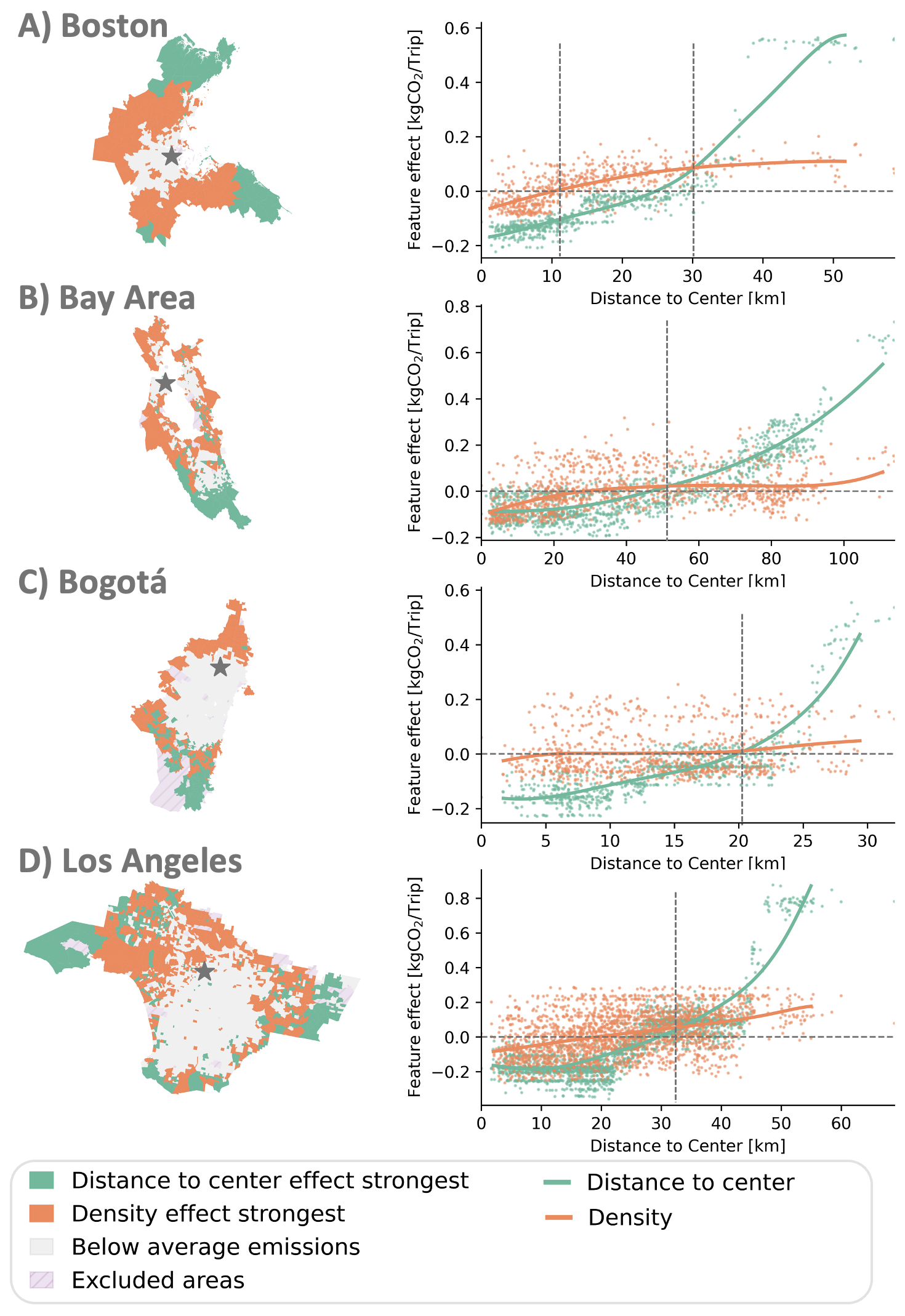}
\caption{\textbf{Threshold  effects in Boston, Bay Area, Bogotá and Los Angeles} For all locations with above mean trip emissions the map on the left visualizes if density (orange) or distance to center (green) are contributing more to trip emissions. The scatter plots on the right highlight the effects of both features over distance to center.} 
\label{si_fig3}
\end{figure}

\end{document}